# SPINEX-Clustering: Similarity-based Predictions with Explainable Neighbors Exploration for Clustering Problems

M.Z. Naser[1,2], Ahmed Z. Naser[3]
[1]School of Civil & Environmental Engineering and Earth Sciences (SCEEES), Clemson University, USA
[2]Artificial Intelligence Research Institute for Science and Engineering (AIRISE), Clemson University, USA
E-mail: mznaser@clemson.edu, Website: www.mznaser.com
[3]Department of Mechanical Engineering, University of Manitoba, Canada, E-mail: a.naser@umanitoba.ca

**Abstract**
This paper presents a novel clustering algorithm from the SPINEX (Similarity-based Predictions with Explainable Neighbors Exploration) algorithmic family. The newly proposed clustering variant leverages the concept of similarity and higher-order interactions across multiple subspaces to group data into clusters. To showcase the merit of SPINEX, a thorough set of benchmarking experiments was carried out against 13 algorithms, namely, Affinity Propagation, Agglomerative, Birch, DBSCAN, Gaussian Mixture, HDBSCAN, K-Means, KMedoids, Mean Shift, MiniBatch K-Means, OPTICS, Spectral Clustering, and Ward Hierarchical. Then, the performance of all algorithms was examined across 51 synthetic and real datasets from various domains, dimensions, and complexities. Furthermore, we present a companion complexity analysis to compare the complexity of SPINEX to that of the aforementioned algorithms. Our results demonstrate that SPINEX can outperform commonly adopted clustering algorithms by ranking within the top-5 best performing algorithms and has moderate complexity. Finally, a demonstration of the explainability capabilities of SPINEX, along with future research needs, is presented.

## 1.0 Introduction

Clustering is a fundamental technique in unsupervised machine learning (ML). This technique aims to partition a dataset into groups (clusters) based on intrinsic similarities among data points (i.e., objects are more similar to each other than those in other groups) [1]. The essence of clustering lies in maximizing intra-cluster similarity while minimizing inter-cluster similarity [2]. This turns into a task that becomes increasingly complex as datasets grow in size and dimensionality. The applications of clustering span various domains, including data mining, pattern recognition, image analysis, bioinformatics, and engineering [3].

Historically, clustering algorithms began with simple heuristic methods like K-Means and hierarchical clustering. Such methods were favored due to their intuitive implementation and interpretation. However, these methods often assume clusters are isotropic and equidistant (which is rarely true in real-world data). Thus, the increase in data complexity has driven researchers to explore algorithms that can adapt to the intrinsic properties of data. Consequently, the field has seen the development of more sophisticated algorithms designed to handle complex data structures that include overlapping clusters with arbitrary shapes and varying densities [4]. For example, density-based clustering methods like DBSCAN provide flexibility by focusing on local rather than global properties, allowing them to discover clusters with arbitrary shapes [5].

An elemental notion of clustering algorithms lies in the concept of similarity (or resemblance), which quantifies the closeness or likeness between pairs of data points. Common similarity measures include Euclidean distance, Manhattan distance, cosine similarity, and Jaccard index [6]. For instance, Euclidean distance is well-suited for continuous data in low-dimensional spaces, while cosine similarity excels in high-dimensional and sparse datasets. Naturally, the choice of similarity measure significantly influences the clustering outcome, as it defines the notion of "closeness" in the data space [7]. As such, such similarity measures can fundamentally shape the clusters formed [8].

In parallel to the notion of similarity, neighboring plays a crucial role in clustering. This is because a neighborhood structure provides valuable information about local data density and connectivity. Such a neighborhood structure involves identifying points that are in close proximity to a given data point [9]. For example, the definition of a neighborhood can vary: it may be based on a fixed radius (as in DBSCAN), a fixed number of nearest neighbors (as in k-nearest neighbors algorithms), or adaptively determined based on data distribution (as in OPTICS) [10]. This can be particularly useful in density-based and hierarchical clustering methods.

Despite the progress made so far, unique challenges continue to hinder the traditional algorithms. Such challenges stem from selecting appropriate parameters and scaling with high-dimensional data [11]. Not only is selecting the number of clusters often required for many clustering algorithms, but such a number remains fixed throughout the analysis. Algorithms like X-means attempt to address this by introducing mechanisms to dynamically estimate the optimal number of clusters [12]. Nevertheless, these approaches still struggle with scalability and parameter sensitivity, which can lead to suboptimal performance on large datasets or datasets with complex structures. Furthermore, traditional clustering algorithms do not account for real-world data being embedded in high-dimensional spaces where the notion of proximity may become distorted [13].

Recent advancements in clustering algorithms have focused on addressing these challenges by developing novel techniques such as random projection and locality-sensitive hashing. These techniques aim to more efficiently approximate nearest neighbor searches in high-dimensional spaces [14]. Moreover, incremental and online clustering algorithms have been developed to handle streaming data and large datasets that do not fit into available memory. Furthermore, ensemble clustering methods, which combine multiple clustering methods or results to produce a more robust partitioning, are gaining momentum [15]. Deep learning approaches have also been used to learn low-dimensional data representations before applying clustering algorithms, potentially improving performance on complex, high-dimensional datasets [16].

From this lens, this paper presents the development of the SPINEX (Similarity-based Predictions with Explainable Neighbors Exploration) clustering algorithm. SPINEX focuses on clustering based on similarity and higher-order interactions across multiple subspaces and incorporates mechanisms for explainability and neighbor exploration. This dual focus enhances the clustering outcomes and provides users with understandable, actionable insights into how clusters are formed, which is crucial for applications where justification of results is required. SPINEX addresses several gaps in existing methodologies by integrating flexibility in the number of clusters, robustness to noise and outliers, and scalability to large datasets. The algorithm's ability to perform well on diverse datasets is demonstrated through extensive benchmarking against 13 clustering algorithms and across 51 synthetic and real datasets.

The rest of the paper is organized as follows: Section 2 describes the main methods and functions comprising SPINEX. Section 3 describes the utilized benchmarking algorithms. Section 4 presents the selected datasets, our benchmarking results, and the outcome of complexity and explainability analyses. The paper concludes with Sections 5 and 6 by highlighting future research directions and our conclusions.

## 2.0 Description of the SPINEX for clustering

*2.1 General description*

SPINEX is an unsupervised clustering algorithm designed to uncover meaningful patterns and groupings in complex datasets. At its core, SPINEX is built on the principle that similarity between data points is key to forming coherent clusters. Unlike traditional clustering algorithms that often rely on a single approach, SPINEX employs several strategies to adapt to various data types and structures. For example, SPINEX utilizes multiple similarity measures, including correlation, Spearman rank correlation, kernel-based similarity, and cosine similarity. This diverse set of measures allows SPINEX to capture different aspects of relationships between data points, making it versatile across various data types and distributions. SPINEX also employs an adaptive approach as it dynamically adjusts its parameters and techniques based on the specific characteristics of the dataset at hand. SPINEX's flexibility extends to its ability to work with or without a predefined number of clusters (i.e., it can autonomously determine an appropriate number of clusters based on the data's structure or work within user-specified constraints).

SPINEX incorporates a multi-level clustering approach, which allows it to uncover hierarchical structures within data. This means it can identify both broad, overarching groups and more nuanced sub-groups. Another key aspect of SPINEX is its focus on explainability, as it offers insights into why certain data points are grouped together. This feature can be valuable in fields where interpretability is important. SPINEX incorporates optimization techniques to handle large datasets, such as dimensionality reduction and parallel processing. SPINEX also assesses the quality of its clusters using various metrics. This self-evaluation helps fine-tune the clustering process.

*2.2 Detailed description*

A more detailed description of SPINEX's functions is provided herein.

### Initialization (__init__)

The __init__ method of the SPINEX initializes the algorithm and sets up its operational parameters. The method signature is as follows:

```
def __init__(self, threshold='auto', n_clusters=None, use_pca=False, n_components=None,
        enable_similarity_analysis=False, enable_neighbor_analysis=False, similarity_methods=None,
        evaluation_tier=1, ground_truth=None, use_approximation=False,
        approximation_method='random_sampling', sample_size=0.5, use_parallel=False,
        parallel_threshold=5000, max_workers=None,use_multi_level=False, multi_level_params=None,
        max_features=100):
```

More specifically:
- threshold: Threshold value for determining cluster membership.
- n_clusters: Number of clusters (if specified).
- use_pca: Boolean to decide whether to use PCA for dimensionality reduction.
- n_components: Number of principal components if PCA is used.
- enable_similarity_analysis: Enables detailed analysis of the similarity data.
- enable_neighbor_analysis: Enables analysis of neighbor relationships in the data.
- similarity_methods: List of methods to compute similarity. Defaults include correlation, spearman rank correlation, kernel methods, and cosine similarity.
- evaluation_tier: Specifies the tier of evaluation metrics used to assess clustering performance.
- use_approximation: Enables the use of approximation methods to speed up computation.
- approximation_method: Specifies the type of approximation method (e.g., 'random_sampling').
- sample_size: Fraction of the data used in the approximation method.
- use_parallel: Enables parallel processing to improve performance.
- parallel_threshold: The minimum size of the dataset to trigger parallel processing.
- max_workers: Maximum number of parallel workers; defaults to system CPU count if parallel processing is enabled.
- use_multi_level: Enables multi-level clustering.
- multi_level_params: Parameters for multi-level clustering (e.g., number of levels, initial threshold).
- max_features: Maximum number of features used in clustering.

## Method: hash_matrix

This method generates a unique hash for a given matrix using SHA-256 to facilitate caching. Let M be a matrix, and the hash H of M is computed as H=hash(M).

## Method: get_similarity_matrix

This method computes or retrieves a cached similarity matrix based on the specified method. These methods span Pearson correlation, Spearman correlation, Kernel matrix ($K(x,y) = e^{-\gamma \|x-y\|^2}$) and Cosine similarity matrix ($C(x,y) = \frac{X \cdot Y}{\text{norm of X and Y}}$).

## Method: log_decision, calculate_correlation_matrix, calculate_spearman_matrix, calculate_kernel_matrix, calculate_cosine_matrix

These methods provide the actual calculations or logging necessary for the algorithm to process data and track decisions.

## Method: apply_approximation

This method reduces the dimensionality or sample size of the data X based on one of the specified approximation methods to expedite the clustering process.

- **Random Sampling**:
    - Randomly selects a subset of data points without replacement.
    - The size of the subset is determined by self.sample_size, a fraction of the total data points.
    - Mathematically: $X_{sampled}=X[I]$, where I is a set of indices randomly chosen such that $|I|=\text{int}(n \times \text{sample\_size})$ and n is the total number of data points.
- **PCA (Principal Component Analysis)**:
    - Reduces the dimensionality by transforming the data into a new coordinate system such that the greatest variance comes to lie on the first few principal axes.
    - Mathematically: $X_{PCA} = PCA(X)$.
- **t-SNE (t-Distributed Stochastic Neighbor Embedding)**:

- o  Projects the data into a lower-dimensional space while trying to maintain the local structure of the data.
- o  Mathematically: $X_{t\text{-}SNE}$ = t-SNE(X).
- **UMAP (Uniform Manifold Approximation and Projection)**:
  - o  A manifold learning technique for dimension reduction.
  - o  Mathematically: $X_{UMAP}$ = UMAP(X).

### Method: set_threshold

This method determines a threshold value to be used in clustering based on the similarity matrix sim_matrix. It offers the following modes of setting this threshold:

**Auto**:
- Automatically sets the threshold based on the statistical properties of the similarity values that are above the median.
- Computes the median of sim_matrix, then identifies values above this median.
- The threshold is set to the median plus one standard deviation of these values: threshold = median + σ.

**Percentile**:
- Sets the threshold at a specific percentile of the similarity values
- Mathematically: threshold = Percentile(sim_matrix).

**Fixed Value:**
- Directly uses a user-specified numeric value as the threshold

### Method: all_similarity_clustering

This method performs clustering on dataset X using all the specified similarity methods in parallel to improve performance:
- If the dataset size exceeds the parallel_threshold and parallel_processing is enabled, it uses a ProcessPoolExecutor to compute the clusters concurrently for each similarity method.
- Each method is applied to dataset X using improved_similarity_clustering, which computes clusters based on the computed similarity matrices.
- Results include the number of clusters and labels for each clustering method.

### Method: calculate_metrics

This method evaluates the quality of clusters obtained from dataset X using a variety of metrics. Here's how it functions:
- **Caching**: It first computes a hash for the data X and the labels to check if the results are already cached to prevent redundant calculations.
- **Metric Calculation**: If not cached, the method calculates several internal and external quality clustering quality metrics:
  - o  Silhouette Score: Measures how similar an object is to its own cluster compared to other clusters. The value ranges from -1 to 1, where higher values indicate better clustering. This metric is calculated as:
    - $$\text{Silhouette} = \frac{b\text{-}a}{\max(a,b)}$$
    - where a is the mean distance to the points in the same cluster, and b is the mean distance to the points in the nearest cluster.
  - o  Calinski-Harabasz Index: Ratio of the sum of between-clusters dispersion and of within-cluster dispersion for all clusters. This metric is calculated as:
    - $$\text{Calinski} - \text{Harabasz} = \frac{\text{Tr}(B_k)}{\text{Tr}(W_k)} \times \frac{N\text{-}k}{k\text{-}1}$$
    - where $B_k$ and $W_k$ are the between and within-cluster scatter matrices, respectively, N is the number of points, and k is the number of clusters.

- o  Davies-Bouldin Index: Defined as the average similarity measure between each cluster and its most similar cluster, where similarity is the ratio of within-cluster distances to between-cluster distances.
    - $\text{Davies} - \text{Bouldin} = \frac{1}{k}\sum_{i=1}^{k} \max_{j \neq i}(\frac{\sigma_i + \sigma_j}{d(c_i, c_j)})$
- o  Homogeneity, Completeness, V-Measure: These are external validation indices that compare the clusters to a ground truth assignment.
    - Homogeneity: Each cluster contains only members of a single class.
        - $h = 1 - H(C)/H(C \mid K)$
        - Where, H(C|K) is the conditional entropy of the classes given the cluster assignments. H(C) is the entropy of the classes.
    - Completeness: All members of a given class are assigned to the same cluster.
        - $c = 1 - H(K \mid C)/H(K)$
        - Where, H(K|C) is the conditional entropy of the clusters given the class labels.. H(K) is the entropy of the clusters.
    - V-Measure: The harmonic mean between homogeneity and completeness.
        - $v = 2\frac{h \cdot c}{h + c}$
        - Where, h is homogeneity and c is completeness.

### Method: find_best_clustering

This method determines the best clustering approach from multiple methods based on calculated metrics:
- Parallel Processing: If enabled and the dataset is large enough, it processes each clustering method in parallel to calculate metrics.
- Score Calculation: A composite_score function evaluates each method based on its metrics and the specified evaluation tier.
- Selection: The method with the highest score is selected as the best.

### Method: apply_pca

This method applies PCA to reduce the dimensionality of the dataset X, improving the efficiency of clustering:
- Standardization: It first standardizes the data using StandardScaler to ensure that PCA works effectively.
- PCA Transformation and Caching: PCA is applied, and the results are either retrieved from the cache or computed and then cached.
- Return: The dimension-reduced data is returned if PCA is enabled; otherwise, the original data is returned.

### Method: improved_similarity_clustering

This method performs clustering based on a similarity matrix derived from the data. It integrates multiple options, including dimensionality reduction via PCA, to enhance clustering based on structural similarity.

- **PCA Application**:
    - o  If PCA is enabled (i.e., self.use_pca is True), it reduces the dimensionality of the data using the apply_pca method to potentially enhance the clarity of clusters by reducing noise and irrelevant features.
- **Similarity Matrix Calculation**:
    - o  It logs the shape of X before similarity matrix calculation, providing a reference for the data's dimensionality at this stage.
    - o  Validates the chosen similarity_method against the supported methods. If an invalid method is specified, it raises a ValueError.

- o   Retrieves or calculates the similarity matrix using get_similarity_matrix, and logs its shape, giving insights into the data's relational structure post-calculation.
- **Threshold Setting**:
  - o   Determines a threshold for clustering based on the calculated similarity matrix using the set_threshold method.
- **Clustering**:
  - o   Performs clustering from the similarity matrix using the cluster_from_similarity method, with the number of data points n and the calculated threshold.
  - o   Returns the cluster labels and the similarity method used, providing a way to evaluate or compare different similarity-based clustering outcomes.

## Method: cluster_from_similarity

This method clusters data based on a similarity matrix and adjusts its approach based on the following configurations:
- **Scalar or Single-Element Matrix**:
  - o   If the similarity matrix is a scalar or contains only one element, it assigns all data points to a single cluster. This handles edge cases where the similarity is uniform or undefined.
- **Multi-Level Clustering**:
  - o   If multi-level clustering is enabled (self.use_multi_level is True), it attempts to cluster the data into multiple levels using predefined parameters such as initial_threshold and levels.
  - o   Logs the completion of multi-level clustering and any errors that occur, defaulting to an optimized merging method if multi-level clustering fails.
- **Hierarchical Clustering**:
  - o   If a specific number of clusters (self.n_clusters) is defined and applicable (i.e., less than the number of data points and the similarity matrix has more than one unique value):
    - ▪   Converts the similarity matrix to a distance matrix where distances are inversely related to similarities.
    - ▪   Applies hierarchical clustering using the linkage method with a complete (maximum) linkage criterion.
    - ▪   Converts the hierarchical cluster structure into flat clusters using a maximum cluster count criterion, adjusting labels to zero-based indexing.
    - ▪   Handles potential errors in hierarchical clustering, defaulting to zero labels if an error occurs.
- **Optimized Merging Method**:
  - o   If no specific clustering mode is applicable, it uses an optimized method for merging clusters based on the provided threshold, which adjusts cluster boundaries dynamically to form cohesive groups.

## Method: should_merge

This method evaluates whether two clusters should be merged based on their similarity. The decision is driven by an external function should_merge_optimized, which considers specific metrics of similarity between the clusters and compares it against a threshold.
- **Parameters**:
  - o   cluster1 and cluster2: Indices or identifiers for the two clusters being considered for merging.
  - o   sim_matrix: A matrix that contains similarity scores between all pairs of observations in the dataset.
  - o   threshold: A cutoff value above which clusters are considered similar enough to merge.
- **Process**:
  - o   The method simply delegates to should_merge_optimized, which implements the actual merging logic based on the similarities provided in sim_matrix and the threshold.

## Method: similarity_contribution_analysis

This method analyzes how each feature contributes to the similarity of a particular observation with all others in the dataset. It is useful for understanding which features are most influential in defining the dataset's similarity (or dissimilarity).

- **Parameters**:
  - X: The dataset matrix where rows are observations and columns are features.
  - observation_index: Index of the observation for which to analyze similarity contributions.
- **Process**:
  - Calculates the correlation matrix for the dataset.
  - Extracts the similarity scores for the specified observation against all others.
  - Computes the absolute differences between the selected observation and all others for each feature.
  - Maps these differences to each feature to understand their contribution to the similarity or dissimilarity.
- **Output**:
  - Returns the similarity scores and a dictionary of contributions per feature, illustrating how each feature's variance from the selected observation impacts the overall similarity.

## Method: nearest_neighbor_analysis

This method identifies the nearest neighbors of a given observation based on similarity scores and analyzes the contribution of each feature to these similarities.

- **Parameters**:
  - X: The dataset matrix.
  - observation_index: Index of the observation for which neighbors are to be found.
  - k: The number of nearest neighbors to identify (default is 5).
- **Process**:
  - Calculates the correlation matrix for X.
  - Extracts the similarity scores for the specified observation.
  - Determines the k nearest neighbors based on these scores.
  - Calculates the absolute differences in feature values between the observation and each of its neighbors to assess how much each feature contributes to making them similar.
- **Output**:
  - Returns a list of the nearest neighbor indices and a dictionary that maps each neighbor index to another dictionary. This nested dictionary breaks down the feature-wise contributions to the similarity, providing detailed insights into what makes these observations close neighbors.

## Method: fit_predict

This method orchestrates the entire clustering process, from data preparation through clustering to explainability analysis:

- **Input Validation and Reshaping**:
  - Ensures that the data X is either 1D or 2D. If X is 1D, it reshapes it to 2D (shape (−1,1)) for consistency.
  - Raises an error if X has more than 2 dimensions.
- **Dimensionality Reduction Using PCA**:
  - If the number of features exceeds self.max_features, PCA is applied to reduce the dimensionality to self.max_features. This is performed to manage complexity and potentially enhance clustering performance.
  - $X_{reduced} = PCA_{max\_features}(X)$
- **Clustering Process**:

- Calls find_best_clustering to identify the best method for clustering and to obtain the initial cluster labels.
- Optionally applies multi-level clustering if enabled, adjusting the labels based on a different, possibly more granular clustering approach:
  - Retrieves the similarity matrix for the best method.
  - Applies cluster_from_similarity without a specific threshold, allowing the method to determine natural cluster divisions.
- **Logging**:
  - Logs the best clustering method used for transparency and traceability.
- **Explainability Analysis**:
  - If either similarity analysis or neighbor analysis is enabled, it processes each observation in X:
    - Parallel Processing: If the number of observations is above the parallel threshold and parallel processing is enabled, it performs analyses concurrently using a ProcessPoolExecutor:
      - Submits tasks for similarity_contribution_analysis and nearest_neighbor_analysis for each observation.
      - Aggregates results and populates self.explainability_results with detailed analyses of how each observation relates to others in terms of feature contributions and nearest neighbors.
    - Sequential Processing: If not using parallel processing, it sequentially processes each observation, storing results in self.explainability_results.
    - Explainability Results Structure:
      - Each observation's results are structured to include both the similarity contributions and nearest neighbor details:
        - ✓ Similarity Analysis: For each feature, show how its variance affects the observation's overall similarity to others.
        - ✓ Neighbor Analysis: Lists the nearest neighbors and details feature contributions that define these neighbor relationships.
- **Return Value**:
  - Returns the final clustering labels derived from the best method and, if applicable, refined by multi-level clustering.

### Method: get_explainability_results

This simple accessor method returns the self.explainability_results dictionary, which holds detailed explainability data for each observation in the dataset.

*Additional functions for optimized clustering:*

### Method: should_merge_optimized

This function determines whether two clusters should be merged based on their average similarity exceeding a given threshold.

- **Parameters**:
  - cluster1, cluster2: Index arrays of the observations in each cluster.
  - sim_matrix: A matrix where the entry at position (i, j) represents the similarity between the i-th and j-th observations.
  - threshold: A scalar value against which the average similarity is compared.
- **Process**:
  - Initializes total_sim to accumulate the total similarity and count to count the number of comparisons.
  - Iterates through all pairs of observations (i, j) where i is in cluster1 and j is in cluster2, adding the similarity to total_sim and incrementing count.

- Returns True if the average similarity (computed as total_sim / count) exceeds the threshold, False otherwise.
- **Mathematical Representation**:
  - merge = $\left(\frac{1}{\text{count}}\sum_{i\in\text{cluster 1}, j\in\text{cluster2}} \text{sim\_matrix}[i,j]\right) >$ threshold

### Method: merge_clusters_optimized

This function iteratively merges clusters based on the should_merge_optimized criteria until no further merges are possible.

- **Parameters**:
  - sim_matrix: Similarity matrix used to evaluate cluster merging.
  - threshold: Similarity threshold for merging clusters.
- **Process**:
  - Initializes clusters with each observation as its own cluster.
  - Iteratively tries to merge clusters:
    - For each pair of clusters (i, j), checks if they should be merged.
    - If a merge is feasible, updates the cluster labels and sets changed to True to continue the merging process in the next iteration.
  - Returns the final cluster labels.

### Method: dynamic_threshold

This function attempts to find a dynamic threshold for clustering by adjusting the initial threshold based on the clustering results.

- **Parameters**:
  - sim_matrix: Similarity matrix for the observations.
  - initial_threshold: Starting threshold for merging.
  - decay_rate: Factor by which the threshold is adjusted.
  - max_iterations: Maximum number of iterations to adjust the threshold.
- **Process**:
  - Adjusts the threshold dynamically, attempting to merge clusters at each new threshold until a stable number of clusters is found or the iteration limit is reached.
  - Returns the adjusted threshold.

### Method: composite_score

This function calculates a composite score for clustering based on selected evaluation metrics.

- **Parameters**:
  - metrics: Dictionary containing clustering evaluation metrics.
  - evaluation_tier: Indicates which metrics to use for scoring.
  - ground_truth: Actual labels, used for external metrics.
- **Process**:
  - Depending on evaluation_tier, calculates a weighted sum of the metrics:
    - Tier 1: Uses internal metrics like Silhouette, Calinski-Harabasz, and Davies-Bouldin.
    - Tier 2: Uses external metrics like Homogeneity, Completeness, and V-Measure.
    - Tier 3: Combines both internal and external metrics.
- **Mathematical Representation**:
  - Score = $\sum w_i \times \text{metric}_i$

### Method: multi_level_clustering

This function implements a hierarchical or multi-tier clustering approach that dynamically adjusts based on changes in cluster structure and inter-cluster similarities across several levels. It is designed to iteratively refine clusters, potentially improving granularity or adjusting to a more natural clustering resolution based on the data's intrinsic properties.

**Parameters**:
- sim_matrix: Initial similarity matrix for the data, where each element indicates how similar each pair of observations is.
- initial_threshold: The starting threshold for merging clusters, which is adapted in subsequent levels.
- levels: Number of iterative levels or tiers of clustering to perform.

**Process**:
- **Initialization**:
    - Sets up an array clusters where each observation initially represents its own cluster.
    - Initializes variables for tracking the number of clusters and variance of the similarity matrix across levels.
    - Prepares a thresholds array filled with the initial threshold, adjusted at each level based on the data's clustering dynamics.
- **Iterative Clustering Across Levels**:
    - For each level:
        - Merges clusters using the merge_clusters_optimized function with the current threshold, producing a set of subclusters.
        - Updates the clustering by mapping old cluster indices to new, more consolidated indices.
        - Breaks out of the loop if no effective change in the number of clusters occurs (indicative of stability or maximal possible merging under the given similarity constraints).
- **Update Similarity Matrix**:
    - Constructs a new similarity matrix for the newly formed clusters by averaging the similarities of observations grouped into the new clusters. This recalculates the relationships based on the current clustering structure, allowing for refined merging in subsequent levels.
    - Efficiently computes this matrix using masks to select relevant subsets of the original matrix.
- **Threshold Adjustment**:
    - Adjusts the thresholds for subsequent levels based on changes in variance and the number of clusters:
        - Reduces thresholds if the variance decreases (clusters becoming more internally consistent).
        - Modifies thresholds based on the rate of change in the number of clusters, tightening the threshold if clusters are changing significantly.
- **Preparation for Next Level**:
    - Updates tracking variables (prev_variance, prev_num_clusters) and sets up the sim_matrix for the next iteration.
    - Translates the current clusters into a form compatible with the newly condensed similarity matrix for subsequent processing.

**Return**:
- The final set of cluster labels after potentially multiple levels of merging and refinement.

**Mathematical Considerations:**
This function utilizes:
- **Averaging of Similarities**: For new clusters i and j, calculates the mean of the relevant elements in the original similarity matrix: new_sim_matrix[i,j] = mean(sim_values). where sim_values are the similarities between observations currently in clusters i and j.
- **Variance Tracking and Adjustment**: Monitors how the variance in cluster similarity changes to dynamically adjust the merging threshold, helping tune the clustering to maintain sensitivity to the data's natural groupings.
- **Cluster Mapping and Re-mapping**: Manages cluster identities through an inverse index mapping to ensure that as clusters are merged, their relationships are maintained and updated correctly.

The complete class of SPINEX is shown below:

```python
class SPINEXClustering:
    def __init__(self, threshold='auto', n_clusters=None, use_pca=False, n_components=None,
                 enable_similarity_analysis=False, enable_neighbor_analysis=False, similarity_methods=None,
                 evaluation_tier=1, ground_truth=None, use_approximation=False,
                 approximation_method='random_sampling', sample_size=0.5, use_parallel=False,
                 parallel_threshold=5000, max_workers=None, use_multi_level=False, multi_level_params=None,
                 max_features=100):
        self.threshold = threshold
        self.similarity_methods = similarity_methods or ['correlation', 'spearman', 'kernel', 'cosine']
        self.use_pca = use_pca
        self.n_components = n_components
        self.enable_similarity_analysis = enable_similarity_analysis
        self.enable_neighbor_analysis = enable_neighbor_analysis
        self.evaluation_tier = evaluation_tier
        self.use_approximation = use_approximation
        self.approximation_method = approximation_method
        self.sample_size = sample_size
        self.decision_log = []
        self.explainability_results = {}
        self.ground_truth = ground_truth.tolist() if isinstance(ground_truth, np.ndarray) else ground_truth
        self.use_parallel = use_parallel
        self.parallel_threshold = parallel_threshold
        self.n_clusters = n_clusters
        self.max_workers = max_workers if max_workers is not None else (mp.cpu_count() if use_parallel else 1)
        self.similarity_cache = {}  # Cache for storing similarity matrices
        self.use_multi_level = use_multi_level
        self.multi_level_params = multi_level_params or {'levels': 3, 'initial_threshold': 0.5}
        self.max_features = max_features

    def hash_matrix(self, matrix):
        """Create a hash for a matrix."""
        m_hash = hashlib.sha256()
        m_hash.update(matrix.data.tobytes())
        return m_hash.hexdigest()

    def get_similarity_matrix(self, X, method):
        """Calculate or retrieve a cached similarity matrix based on the method."""
        if X.shape[1] < 2 and (method == 'correlation' or method == 'spearman'):
            similarity_matrix = np.ones((X.shape[0], X.shape[0]))  # Perfect similarity
        else:
            matrix_hash = self.hash_matrix(X)
            cache_key = (matrix_hash, method)
            if cache_key in self.similarity_cache:
                similarity_matrix = self.similarity_cache[cache_key]
```

```python
                self.log_decision(f"Retrieved {method} similarity matrix from cache.")
            else:
                if method == 'correlation':
                    similarity_matrix = self.calculate_correlation_matrix(X)
                elif method == 'spearman':
                    similarity_matrix = self.calculate_spearman_matrix(X)
                elif method == 'kernel':
                    similarity_matrix = self.calculate_kernel_matrix(X)
                elif method == 'cosine':
                    similarity_matrix = self.calculate_cosine_matrix(X)
                else:
                    raise ValueError(f"Invalid similarity method: {method}")
                self.similarity_cache[cache_key] = similarity_matrix
                self.log_decision(f"Computed and cached {method} similarity matrix.")
        # Ensure the similarity matrix is 2D
        if similarity_matrix.ndim == 0:
            similarity_matrix = np.array([[similarity_matrix]])
        elif similarity_matrix.ndim == 1:
            similarity_matrix = similarity_matrix.reshape(1, -1)
        return similarity_matrix

    def log_decision(self, message):
        self.decision_log.append(message)

    @staticmethod
    def calculate_correlation_matrix(X):
        if X.shape[1] < 2:
            return np.ones((X.shape[0], X.shape[0]))
        return np.corrcoef(X)

    @staticmethod
    def calculate_spearman_matrix(X):
        if X.shape[1] < 2:
            return np.ones((X.shape[0], X.shape[0]))
        return spearmanr(X)[0]

    @staticmethod
    def calculate_kernel_matrix(X, gamma=1.0):
        return rbf_kernel(X, gamma=gamma)

    @staticmethod
    def calculate_cosine_matrix(X):
        return cosine_similarity(X)

    def apply_approximation(self, X):
```

```python
        if self.use_approximation:
            if self.approximation_method == 'random_sampling':
                sample_indices = np.random.choice(X.shape[0], int(X.shape[0] * self.sample_size), replace=False)
                X = X[sample_indices]
                self.log_decision(f"Data reduced to {X.shape[0]} samples using random sampling.")
            elif self.approximation_method == 'pca':
                pca = PCA(n_components=self.n_components if self.n_components is not None else 0.95)
                X = pca.fit_transform(X)
                self.log_decision(f"Data reduced to {pca.n_components_} dimensions using PCA.")
            elif self.approximation_method == 'tsne':
                tsne = TSNE(n_components=self.n_components if self.n_components is not None else 2)
                X = tsne.fit_transform(X)
                self.log_decision("Data projected to lower dimensions using t-SNE.")
            elif self.approximation_method == 'umap':
                umap = UMAP(n_components=self.n_components if self.n_components is not None else 2)
                X = umap.fit_transform(X)
                self.log_decision("Data projected to lower dimensions using UMAP.")
        return X

    def set_threshold(self, sim_matrix):
        if self.threshold == 'auto':
            median_val = np.median(sim_matrix)
            above_median = sim_matrix[sim_matrix > median_val]
            if len(above_median) > 0:
                std_dev = np.std(above_median)
                threshold_value = median_val + std_dev
            else:
                threshold_value = np.max(sim_matrix)
            self.log_decision(f"Adaptive threshold set using density-based approach: {threshold_value}")
        elif isinstance(self.threshold, str) and self.threshold.endswith('%'):
            percentile = float(self.threshold[:-1])
            threshold_value = np.percentile(sim_matrix.flatten(), percentile)
            self.log_decision(f"Threshold set using percentile: {threshold_value}")
        elif isinstance(self.threshold, (int, float)):
            threshold_value = self.threshold
            self.log_decision(f"Threshold set using fixed value: {threshold_value}")
        else:
            raise ValueError("Invalid threshold specified")
        return threshold_value

    def all_similarity_clustering(self, X):
        if self.use_parallel and X.shape[0] >= self.parallel_threshold:
            with ProcessPoolExecutor(max_workers=self.max_workers) as executor:
                futures = {method: executor.submit(self.improved_similarity_clustering, X, method)
                           for method in self.similarity_methods}
```

```python
            results = {method: future.result() for method, future in futures.items()}
    else:
        results = {method: self.improved_similarity_clustering(X, method)
                   for method in self.similarity_methods}
    return {method: {'n_clusters': len(np.unique(labels)), 'labels': labels}
            for method, (labels, _) in results.items()}

def calculate_metrics(self, X, labels, method):
    # Create a hash for the data and labels to serve as a cache key
    data_hash = self.hash_matrix(X)
    labels_hash = self.hash_matrix(labels)
    cache_key = (data_hash, labels_hash, method)
    # Check if metrics are already calculated and stored in cache
    if cache_key in self.similarity_cache:
        self.log_decision(f"Metrics retrieved from cache for method: {method}")
        return self.similarity_cache[cache_key]
    # Compute metrics if not in cache
    n_clusters = len(np.unique(labels))
    metrics = {
        'n_clusters': n_clusters,
        'labels': labels,
        'Silhouette': np.nan,
        'Calinski-Harabasz': np.nan,
        'Davies-Bouldin': np.nan,
        'Homogeneity': np.nan,
        'Completeness': np.nan,
        'V-Measure': np.nan
    }
    if 1 < n_clusters < len(X):
        try:
            if self.evaluation_tier in [1, 3]:
                metrics['Silhouette'] = silhouette_score(X, labels)
                metrics['Calinski-Harabasz'] = calinski_harabasz_score(X, labels)
                metrics['Davies-Bouldin'] = davies_bouldin_score(X, labels)
            if self.evaluation_tier in [2, 3] and self.ground_truth is not None:
                metrics['Homogeneity'] = homogeneity_score(self.ground_truth, labels)
                metrics['Completeness'] = completeness_score(self.ground_truth, labels)
                metrics['V-Measure'] = v_measure_score(self.ground_truth, labels)
        except Exception as e:
            self.log_decision(f"Error calculating metrics for {method}: {str(e)}")
    # Cache the calculated metrics
    self.similarity_cache[cache_key] = metrics
    self.log_decision(f"Metrics computed and cached for method: {method}")
    return metrics
```

```python
def find_best_clustering(self, X):
    results = self.all_similarity_clustering(X)
    if len(results) == 1:
        best_method = list(results.keys())[0]
        return results[best_method]['labels'], best_method
    if self.use_parallel and X.shape[0] >= self.parallel_threshold:
        with ProcessPoolExecutor(max_workers=self.max_workers) as executor:
            futures = {method: executor.submit(self.calculate_metrics, X, results[method]['labels'], method)
                       for method in results}
            detailed_results = {method: future.result() for method, future in futures.items()}
    else:
        detailed_results = {method: self.calculate_metrics(X, results[method]['labels'], method)
                            for method in results}
    scored_results = {method: composite_score(metrics, self.evaluation_tier, self.ground_truth)
                      for method, metrics in detailed_results.items()}
    best_method = max(scored_results, key=scored_results.get)
    return detailed_results[best_method]['labels'], best_method

def apply_pca(self, X):
    if self.use_pca:
        scaler = StandardScaler()
        X_scaled = scaler.fit_transform(X)
        matrix_hash = self.hash_matrix(X_scaled)  # Hash the scaled data
        cache_key = ('PCA', matrix_hash)
        if cache_key in self.similarity_cache:
            pca_result = self.similarity_cache[cache_key]
            self.log_decision("Retrieved PCA results from cache.")
        else:
            pca = PCA(n_components=self.n_components if self.n_components is not None else 0.95)
            pca_result = pca.fit_transform(X_scaled)
            self.similarity_cache[cache_key] = pca_result  # Cache the PCA result
            self.log_decision(f"Computed and cached PCA results. Reduced dimensions to {pca.n_components_}.")
        return pca_result
    return X

def improved_similarity_clustering(self, X, similarity_method='correlation'):
    if self.use_pca:
        X = self.apply_pca(X)
    self.log_decision(f"Shape of X before similarity matrix calculation: {X.shape}")
    if similarity_method not in self.similarity_methods:
        raise ValueError(f"Invalid similarity method. Choose from {', '.join(self.similarity_methods)}.")
    sim_matrix = self.get_similarity_matrix(X, similarity_method)
    self.log_decision(f"Similarity matrix shape: {sim_matrix.shape}")
    threshold = self.set_threshold(sim_matrix)
    labels = self.cluster_from_similarity(sim_matrix, X.shape[0], threshold)
```

```python
        return labels, similarity_method

    def cluster_from_similarity(self, sim_matrix, n, threshold):
        if np.isscalar(sim_matrix) or sim_matrix.size == 1:
            return np.zeros(n, dtype=int)  # Return all points in one cluster
        if self.use_multi_level:
            self.log_decision("Using multi-level clustering")
            try:
                clusters = multi_level_clustering(sim_matrix,
                                    self.multi_level_params['initial_threshold'],
                                    levels=self.multi_level_params['levels'])
                self.log_decision(f"Multi-level clustering completed with {len(np.unique(clusters))} clusters")
            except Exception as e:
                self.log_decision(f"Error in multi-level clustering: {str(e)}. Falling back to default clustering.")
                clusters = merge_clusters_optimized(sim_matrix, float(threshold) if threshold is not None else 0.5)
            return clusters
        if self.n_clusters is not None and self.n_clusters < n and sim_matrix.size > 1:
            # Use hierarchical clustering when n_clusters is set
            distance_matrix = 1 - np.clip(sim_matrix, -1, 1)
            np.fill_diagonal(distance_matrix, 0)
            distance_matrix = (distance_matrix + distance_matrix.T) / 2
            distance_matrix = np.maximum(distance_matrix, 0)
            try:
                linkage_matrix = linkage(squareform(distance_matrix), method='complete')
                cluster_labels = fcluster(linkage_matrix, t=self.n_clusters, criterion='maxclust')
                return cluster_labels - 1
            except ValueError as e:
                print(f"Error in hierarchical clustering: {e}")
                return np.zeros(n)
        else:
            # Use optimized merging method
            clusters = merge_clusters_optimized(sim_matrix, float(threshold) if threshold is not None else 0.5)
            return clusters

    def should_merge(self, cluster1, cluster2, sim_matrix, threshold):
        return should_merge_optimized(cluster1, cluster2, sim_matrix, threshold)

    def get_decision_log(self):
        return self.decision_log

    def similarity_contribution_analysis(self, X, observation_index):
        n_features = X.shape[1]
        sim_matrix = self.calculate_correlation_matrix(X)
        observation_similarities = sim_matrix[observation_index]
        obs_value = X[observation_index]
```

```python
        differences = np.abs(X - obs_value)
        contributions = {f'Feature_{i}': differences[:, i] for i in range(n_features)}
        return observation_similarities, contributions

    def nearest_neighbor_analysis(self, X, observation_index, k=5):
        sim_matrix = self.get_similarity_matrix(X, self.similarity_methods[0])
        observation_similarities = sim_matrix[observation_index]
        nearest_neighbors = np.argsort(observation_similarities)[::-1][1:k+1]
        obs_value = X[observation_index]
        neighbor_values = X[nearest_neighbors]
        contributions = np.abs(neighbor_values - obs_value)
        neighbor_contributions = {
            int(neighbor_index): {f'Feature_{i}': contributions[j, i] for i in range(X.shape[1])}
            for j, neighbor_index in enumerate(nearest_neighbors)
        }
        return nearest_neighbors.tolist(), neighbor_contributions

    def fit_predict(self, X):
        if X.ndim == 1:
            X = X.reshape(-1, 1)
        elif X.ndim > 2:
            raise ValueError("Input array X should be 1D or 2D.")
        # If there are too many features, use PCA to reduce dimensionality
        if X.shape[1] > self.max_features:
            self.log_decision(f"Reducing features from {X.shape[1]} to {self.max_features} using PCA")
            pca = PCA(n_components=self.max_features)
            X = pca.fit_transform(X)
        labels, best_method = self.find_best_clustering(X)
        if self.use_multi_level:
            self.log_decision("Applying multi-level clustering to best similarity matrix")
            sim_matrix = self.get_similarity_matrix(X, best_method)
            labels = self.cluster_from_similarity(sim_matrix, X.shape[0], None)
        self.log_decision(f"Best clustering method: {best_method}")
        if self.enable_similarity_analysis or self.enable_neighbor_analysis:
            if self.use_parallel and X.shape[0] >= self.parallel_threshold:
                with ProcessPoolExecutor(max_workers=self.max_workers) as executor:
                    futures = []
                    for i in range(X.shape[0]):
                        if self.enable_similarity_analysis:
                            futures.append(executor.submit(self.similarity_contribution_analysis, X, i))
                        if self.enable_neighbor_analysis:
                            futures.append(executor.submit(self.nearest_neighbor_analysis, X, i, k=5))
                    results = [future.result() for future in futures]
                    results_index = 0
                    for i in range(X.shape[0]):
```

```
                self.explainability_results[i] = {}
                if self.enable_similarity_analysis:
                    similarities, contributions = results[results_index]
                    self.explainability_results[i]['similarity_analysis'] = {
                        'similarities': similarities.tolist(),
                        'contributions': {k: v.tolist() for k, v in contributions.items()}
                    }
                    results_index += 1
                if self.enable_neighbor_analysis:
                    nearest_neighbors, neighbor_contributions = results[results_index]
                    self.explainability_results[i]['neighbor_analysis'] = {
                        'nearest_neighbors': nearest_neighbors,
                        'neighbor_contributions': neighbor_contributions
                    }
                    results_index += 1
        else:
            for i in range(X.shape[0]):
                self.explainability_results[i] = {}
                if self.enable_similarity_analysis:
                    similarities, contributions = self.similarity_contribution_analysis(X, i)
                    self.explainability_results[i]['similarity_analysis'] = {
                        'similarities': similarities.tolist(),
                        'contributions': {k: v.tolist() for k, v in contributions.items()}
                    }
                if self.enable_neighbor_analysis:
                    nearest_neighbors, neighbor_contributions = self.nearest_neighbor_analysis(X, i, k=5)
                    self.explainability_results[i]['neighbor_analysis'] = {
                        'nearest_neighbors': nearest_neighbors,
                        'neighbor_contributions': neighbor_contributions
                    }
        return labels
    def get_explainability_results(self):
        return self.explainability_results
```

## 3.0 Description of benchmarking algorithms, experiments, and functions

As mentioned above, SPINEX was examined against 13 commonly used clustering algorithms, namely, K-Means, DBSCAN, Agglomerative, Spectral Clustering, Mean Shift, OPTICS, Gaussian Mixture, Birch, Affinity Propagation, MiniBatch K-Means, Ward Hierarchical, HDBSCAN, and KMedoids. Each of these algorithms is described herein, and we note that more details on each algorithm can be found in the cited sources. Table XXX compares these algorithms with respect to SPINEX.

*3.1 Affinity Propagation*
Affinity Propagation was introduced by Frey and Dueck in 2007 [17]. The main logic behind this algorithm stems from its ability to identify exemplars among data points and form clusters based on these exemplars. Thus, Affinity Propagation determines the number of clusters based on data characteristics and input preferences, which makes it suitable for scenarios where the number of clusters is not known a priori. The algorithm operates by sending messages between pairs of data points. There are two types of messages: `responsibility` messages, sent from data points to potential exemplars that reflect the suitability of one data point to be the exemplar of another, and `availability` messages, sent from potential exemplars to data points to indicate how appropriate it would be for a data point to choose the exemplar. The messages are updated iteratively based on the similarity between data points, typically measured as negative squared Euclidean distance. Affinity Propagation can quickly converge to high-quality clusters. The main parameter to adjust is the `preference`, which controls how many exemplars are used, with a higher preference leading to more clusters. This flexibility makes the algorithm versatile for a wide range of clustering tasks. However, the computational complexity rooted in the number of messages that need to be passed increases quadratically with the number of data points, which can be a limitation in large datasets [18]. Moreover, the choice of the similarity measure and the `preference` parameter can significantly affect the clustering outcome.

*3.2 Agglomerative*

Agglomerative clustering is a hierarchical clustering method developed by various researchers, with significant contributions from Ward Jr. in 1963 [19]. This bottom-up approach starts with each data point as a single cluster and iteratively merges the closest clusters until the desired number of clusters is reached. The algorithm calculates the distance between all pairs of clusters, merges the two closest clusters, and repeats this process until the desired number of clusters is reached or only one cluster remains. The agglomerative clustering process can be visualized using a dendrogram, a tree-like diagram showing the sequence of cluster mergers, and the distance at which each merger occurred. This hierarchical approach allows one to interpret the data structure at different clustering scales and offers an insightful visualization of how clusters are related. Some of the primary advantages of agglomerative clustering include its flexibility regarding cluster shape and size, which does not require clusters to be similar. However, one of its main drawbacks is scalability; because the algorithm requires the computation of the distance matrix between all pairs of samples, it can be computationally intensive for large datasets [20].

*3.3 Balanced Iterative Reducing and Clustering using Hierarchies (BIRCH)*

BIRCH was developed by Zhang et al. in 1996 [21]. This algorithm is designed for clustering large datasets, particularly those that do not fit into memory. As such, this algorithm becomes especially effective for applications that require incrementally processing incoming, potentially multi-dimensional data. Birch incrementally constructs a tree-like structure called the Clustering Feature Tree (CF Tree), which summarizes the data points and maintains enough statistical information to execute clustering decisions incrementally. More specifically, the CF Tree efficiently condenses data, and the algorithm only needs a single scan of the data, making it highly scalable and faster than traditional algorithms. This is particularly beneficial in real-time clustering scenarios. BIRCH can be most suitable for spherical clusters and may not perform well with clusters of arbitrary shape.

*3.4 Density-Based Spatial Clustering of Applications with Noise (DBSCAN)*

The DBSCAN was introduced by Ester et al. in 1996 [22]. This density-based clustering algorithm groups together points that are closely packed together. Unlike K-Means, DBSCAN does not require the user to specify the number of clusters beforehand. Instead, DBSCAN operates based on a density estimation of the dataset, where clusters are defined as high-density areas separated by low-density regions. The primary mechanism of DBSCAN involves two parameters: eps (epsilon), which determines the radius around each data point to search for neighboring points, and min_samples, indicating the minimum number of points required to form a dense region. A core point, which has at least min_samples within its eps neighborhood, forms the start of a cluster. From this core point, the cluster is expanded recursively, including reachable density-connected points to allow DBSCAN to naturally grow clusters of arbitrary shape. DBSCAN is particularly useful in environments where data contains outliers or noise. Moreover, the algorithm's intrinsic property of adjusting to varying densities enables effective clustering of data with uneven cluster sizes and densities, which are common in real-world data. However, this algorithm can struggle with datasets of varying densities and high-dimensional spaces and heavily relies on the selected initial parameters [23].

*3.5 Gaussian Mixture*

The Gaussian Mixture Model (GMM) for clustering has its roots in the work of various statisticians and was popularized in the ML community in the 1990s [24]. This probabilistic algorithm assumes all the data points are generated from a mixture of a finite number of Gaussian distributions with unknown parameters. The algorithm uses the Expectation-Maximization (EM) algorithm to fit the Gaussian mixture to the data, iteratively refining the parameters to maximize the likelihood of the observed data. As an extension of the single Gaussian distribution model, the GMM provides a method for soft clustering, where each data point belongs to each cluster to a different degree, as defined by the probability of belonging to each Gaussian component. GMM is capable of modeling clusters that have different sizes and shapes. Furthermore, the soft clustering approach allows for more nuanced interpretations and insights into data categorizations, where data points can exhibit membership in multiple clusters to varying degrees.

*3.6 Hierarchical DBSCAN (HDBSCAN)*

HDBSCAN was developed by Campello et al. in 2013 as an extension of DBSCAN [25]. This variant converts the DBSCAN into a hierarchical clustering algorithm. HDBSCAN relies on the concept of minimum cluster size to improve how clusters are extracted from varying density data. This makes HDBSCAN more flexible and applicable to a wider range of data types and structures. Moreover, the algorithm determines optimal clustering based on stability over different density scales, which helps identify clusters that are consistent across scale changes. However, HDBSCAN can also be computationally intensive due to its hierarchical approach and the requirement to compute and compare stability across different clusters. Additionally, selecting the appropriate parameters like min_cluster_size (which helps to distinguish between noise and clusters effectively) [26].

*3.7 K-Means*

The K-Means algorithm was developed by Lloyd in 1957 [27] (though published much later in 1982). This algorithm aims to partition n observations into k clusters, where each observation belongs to the cluster with the nearest mean (centroid). The K-Means works iteratively by initializing k centroids randomly in the feature space and then assigning each data point to the nearest centroid. Then, the algorithm recalculates the centroids as the mean of all points assigned to that centroid and continues to repeat these steps until the centroids no longer move significantly. K-Means is particularly effective for datasets with spherical clusters of similar size. However, this algorithm may struggle with clusters of varying sizes and densities and is sensitive to the initial placement of centroids – thu much more modern and faster versions now exist [28].

*3.8 KMedoids*
K-Medoids, also known as PAM (Partitioning Around Medoids), was introduced by Kaufman and Rousseeuw in 1987 [29]. This is another variation of K-Means that uses actual data points (medoids) as the center of clusters instead of centroids. This algorithm aims to minimize the sum of dissimilarities between points labeled to be in a cluster and the point designated as the center of that cluster. Thus, the K-Medoids can be more robust to noise and outliers than K-Means, as it minimize a sum of pairwise dissimilarities instead of squared Euclidean distances. The algorithm works iteratively: first, it randomly selects k medoids. Then, each point is assigned to the closest medoid, and the total cost (sum of distances between points and their medoid) is calculated. The algorithm then tries to improve the clustering by swapping each medoid with a non-medoid point and checking if the total cost decreases. This process repeats until no improvements are found. K-Medoids can be computationally more expensive than K-Means, particularly as the number of data points and clusters increases. Additionally, like K-Means, it requires specifying the number of clusters in advance, which can be a drawback if the optimal number of clusters is not known [30].

*3.9 Mean Shift*
Fukunaga and Hostetler introduced the Mean Shift algorithm in 1975 [31] as a non-parametric clustering technique based on finding modes in a smooth density estimate of the input space. Unlike K-Means or hierarchical clustering, Mean Shift does not require the number of clusters to be specified in advance and hence becomes an attractive option for applications where the number of clusters is not known. The algorithm updates candidates for centroids to be the mean of the points within a given region. These regions are defined by a `bandwidth parameter`, which can significantly influence the clustering outcome. Each iteration of the algorithm shifts the centroids toward the peaks of the dataset's density function until convergence, where no centroid moves significantly. It is worth noting that the `bandwidth parameter` determines the radius of the area considered around each data point at each step of the iteration. A smaller bandwidth can lead to a larger number of clusters, whereas a larger bandwidth can merge distinct clusters into a single one. Thus, the choice of bandwidth is crucial and can often be determined using a heuristic based on the data distribution. One of the primary advantages of Mean Shift is its ability to model complex clusters without assuming any prior shape, like spheres or ellipsoids, as is common in other clustering algorithms listed herein. Furthermore, the algorithm's iterative shifting to the denser areas allows it to be robust against outliers and noise [32].

*3.10 MiniBatch K-Means*
MiniBatch K-Means was proposed by David Sculley in 2010 [33] as a variant of the K-Means algorithm. This new variant was designed to be used with large datasets. MiniBatch K-Means improves upon the computational efficiency of K-Means by using small, random subsets of the data (mini-batches) for each algorithm iteration (as opposed to the full dataset). This approach significantly reduces the computation needed to converge to a solution and, in turn, becomes well-suited for scenarios involving very large datasets or streaming data. While MiniBatch K-Means converges faster than K-Means, it may negatively reflect the quality of the results. MiniBatch K-Means can also be sensitive to the choice of the mini-batch size. For example, smaller batches can lead to a higher variance in the results and may require more iterations to converge, potentially offsetting the efficiency gains. Additionally, like K-Means, it assumes clusters to be spherical and may perform poorly on complex geometric shapes or varying cluster sizes [34].

*3.11 Ordering Points To Identify the Clustering Structure (OPTICS)*
The OPTICS algorithm was proposed by Ankerst et al. in 1999 [35]. OPTICS extends the concepts of DBSCAN to overcome its sensitivity to the parameter settings, particularly regarding the choice of the eps parameter. Like DBSCAN, OPTICS deals with density-based spatial clustering, but it introduces a more flexible approach by creating an ordered reachability plot to provide a detailed representation of the dataset's structure at all scales. The core idea of OPTICS revolves around the concept of core-distance and reachability-distance, which are used to assess the density connectivity between points in the dataset. The algorithm processes each point in the dataset, computing these distances and building an ordered list of points representing the density clustering structure. This ordering allows clusters of varying densities to be identified (which can be difficult using DBSCAN as it uses a single eps value, which may not suit all clusters). OPTICS can perform well over datasets with complex structures and varying densities without the need for input parameters related to the distance threshold. This makes it more robust than DBSCAN in scenarios where the density contrast between clusters varies significantly. However, OPTICS can be computationally intensive as it's logic involves complex distance computations [36].

*3.12 Spectral Clustering*
Spectral clustering was popularized by Ng et al. in 2002 [37]. The origin of this algorithm stems from graph theory. The algorithm treats clustering as a graph partitioning problem, where data points are considered as graph nodes. The algorithm connects nodes with edges weighted by the similarity between points, and the goal is to divide this graph so that the edges between different groups have low weights (low similarity) while the edges within a group have high weights (high similarity). More specifically, the algorithm initiates by constructing a similarity matrix using the Gaussian kernel to measure the similarity between data points. Then, it derives the Laplacian matrix to capture the structure of the graph. The next step involves solving an eigenvalue problem for the Laplacian matrix to obtain the eigenvalues and their corresponding eigenvectors. The eigenvectors corresponding to the smallest eigenvalues (except for the smallest, zero) are then used to transform the high-dimensional data into a lower-dimensional space where the clusters become more apparent and can be easily separated using a simple algorithm like K-Means. Spectral clustering can identify complex cluster structures that are not necessarily compact or convex. However, it can be computationally intensive for large datasets [38].

*3.13 Ward Hierarchical*

Ward's method is a specific approach to agglomerative clustering that was developed by Ward Jr. in 1963. The method is particularly focused on minimizing the total within-cluster variance. At each step, the pair of clusters that leads to the minimum increase in total within-cluster variance after merging is chosen. Thus, this method tends to create clusters of similar size and is less susceptible to noise and outliers compared to other hierarchical clustering methods. This makes this method suitable for scenarios where cluster homogeneity is important [39]. However, like other hierarchical clustering methods, Ward's method can be computationally intensive, especially with large datasets, as it requires calculating and updating the distance matrix between all clusters at each iteration. Additionally, because it tends to create clusters of roughly equal size, it may not perform well on data where true clusters vary significantly in size.

Table XXX A comparison among the examined algorithms

| Algorithm | Method Type | Suitable for Large Datasets | Noise Handling | Shape of Clusters | Initialization Dependency |
|---|---|---|---|---|---|
| Affinity Propagation | Exemplar-based | No | Good | Arbitrary | Low |
| Agglomerative | Hierarchical | No | Moderate | Arbitrary | Low |
| Birch | Hierarchical | Yes | Poor | Spherical and non-spherical | Low |
| DBSCAN | Density-based | Moderate | Excellent | Arbitrary | Low |
| Gaussian Mixture | Probabilistic | No | Moderate | Elliptical | Moderate |
| HDBSCAN | Density-based | Moderate | Excellent | Arbitrary | Low |
| K-Means | Partitioning | Yes | Poor | Spherical | High |
| KMedoids | Partitioning | Moderate | Excellent | Arbitrary | Moderate |
| Mean Shift | Mode-seeking | No | Good | Arbitrary | Low |
| MiniBatch K-Means | Partitioning | Yes | Poor | Spherical | High |
| OPTICS | Density-based | Moderate | Excellent | Arbitrary | Low |
| Spectral Clustering | Graph-based | No | Moderate | Arbitrary | Moderate |
| SPINEX | Similarity-based | Yes | Excellent | Arbitrary | Low |
| Ward Hierarchical | Hierarchical | No | Moderate | Compact | Low |

**4.0 Description of benchmarking experiments, algorithms, and datasets**

Our benchmarking analysis comprises a collection of 51 synthetic and real datasets. The complete analysis was run and evaluated in a Python 3.10.5 environment using an Intel(R) Core(TM) i7-9700F CPU @ 3.00GHz and an installed RAM of 32.0GB. To ensure reproducibility, the settings of SPINEX and the other algorithms presented earlier will be found in our Python script, wherein the comparative algorithms ran in default settings, and various versions of SPINEX were examined. The performance and quality of the clustering analysis were evaluated through a number of internal and external performance metrics – see Table XXX [3].

Internal metrics (Do not require ground truth labels):
- Silhouette Score is a measure that balances both cohesion and separation of clusters and ranges from -1 to 1. For each data point, the average distance to points in its own cluster (a) is compared with the average distance to points in the nearest neighboring cluster (b). A high Silhouette score suggests that points are well-matched to their own clusters and poorly-matched to neighboring clusters. This metric is particularly useful when the number of clusters is not known a priori, as it can help identify the optimal number of clusters.

- The Calinski-Harabasz Index evaluates cluster validity based on the average between- and within-cluster sum of squares. This metric calculates the ratio of the between-cluster dispersion mean to the within-cluster dispersion, adjusted for the number of clusters and data points. Higher values indicate better-defined clusters. This index can be particularly effective for hyper-spherical clusters and tends to be higher for convex clusters.
- The Davies-Bouldin Index measures the average similarity between each cluster and its most similar cluster. This index is calculated as the ratio of within-cluster distances to between-cluster distances (with lower values indicating better clustering, as this implies that clusters are compact and well-separated). This index is particularly useful when clusters are expected to be convex and well-separated, and it does not assume any specific number of clusters.

External metrics (require ground truth labels):
- Homogeneity measures the extent to which each cluster contains only members of a single class. This metric is based on the conditional entropy of the class distribution given the cluster assignments. A score of 1 indicates perfect homogeneity, where each cluster contains only members of a single class. This metric is particularly useful when the goal is to ensure that clusters do not mix different classes.
- Completeness measures the extent to which all members of a given class are assigned to the same cluster. It is based on the conditional entropy of the cluster assignments given the class distribution. A score of 1 indicates perfect completeness, where all members of each class are assigned to a single cluster.
- V-Measure is the harmonic mean of homogeneity and completeness. It provides a single score that balances these two often competing criteria. A high V-Measure indicates that the clustering solution has high homogeneity and completeness.

Table XXX List of common performance metrics.

| Metric | Formula |
|---|---|
| Silhouette | $$\text{Silhouette} = \frac{b - a}{max(a, b)}$$ where a is the mean distance to the points in the same cluster, and b is the mean distance to the points in the nearest cluster. |
| Calinski-Harabasz | $$\text{Calinski} - \text{Harabasz} = \frac{Tr(B_k)}{Tr(W_k)} \times \frac{N-k}{k-1}$$ where $B_k$ and $W_k$ are the between and within-cluster scatter matrices, respectively, N is the number of points, and k is the number of clusters. |
| Davies-Bouldin | $$\text{Davies} - \text{Bouldin} = \frac{1}{k}\sum_{i=1}^{k} \max_{j \neq i}(\frac{\sigma_i + \sigma_j}{d(c_i, c_j)})$$ |
| Homogeneity | $$h = 1 - H(C)/H(C \mid K)$$ Where, H(C|K) is the conditional entropy of the classes given the cluster assignments. H(C) is the entropy of the classes. |
| Completeness | $$c = 1 - H(K \mid C)/H(K)$$ Where, H(K|C) is the conditional entropy of the clusters given the class labels. H(K) is the entropy of the clusters. |

| | |
|---|---|
| V-Measure | $$v = 2\frac{h \cdot c}{h + c}$$ Where, h is homogeneity and c is completeness. |

*4.1 Synthetic datasets*

Thirty three synthetic datasets that simulate different clustering conditions and scenarios were generated and used in this leg of experiments. These datasets are described herein as well as in Fig. XXX and Table XXX.

The datasets are generated using several utility functions primarily from the sklearn.datasets module. Each utility function is designed to produce specific types of data structures to evaluate the clustering algorithms in this study. The following is an overview of these functions and their roles:

- make_blobs: Generates isotropic Gaussian blobs for clustering. This function is versatile and can create clusters with varying densities and sizes by adjusting the standard deviation, number of features, and centers. This function is used in the following datasets: Blobs, Anisotropic, Varied Density, Stretched Blobs, Aggregated Clusters, Nested Clusters, Gaussian Mixture, Hierarchical Clusters, Simple Blobs, Shifting Variance Clusters, Disjoint Clusters, and Non-spherical Gaussian Mixture.
- make_moons: Creates a two-dimensional binary dataset with two interleaving half circles, which can form a shape similar to two moons. This dataset can be useful for testing non-linear boundaries and appears in the Moons and Interlocking Moons datasets.
- make_circles: Produces a large circle containing a smaller circle in 2D. This data structure can be ideal for testing algorithms on datasets with clear but complex hierarchical groupings. This function is utilized in Circles, Overlapping Circles, and Concentric Spheres.
- make_swiss_roll: This function generates a Swiss Roll dataset and Manifold Learning dataset.
- Random Data Generators (np.random.rand, np.random.randn, etc.). These functions are from the NumPy library generate random data. np.random.rand creates data uniformly distributed over [0, 1), and np.random.randn generates samples from the standard normal distribution. They are used to create datasets without inherent clustering structures or to add noise. These functions are used in No structure, Random Uniform Scatter, Random Walk Clusters, Sparse High-Dimensional, and Highly Correlated Features.
- Some datasets require specific patterns or structures not readily available in standard libraries. These include Spirals, Checkerboard, Sine Wave Clusters, Broken Rings, Winding Function Clusters, Feature Entanglement, and Periodic Patterns.
- Custom Composite Datasets: These are datasets constructed by combining features or elements from other datasets, often modified by mathematical functions or operations to create complex interactions between features. These are seen in Friedman's Function and others, where complex data interactions are necessary for advanced regression or clustering tasks.

Table XXX Parameters used in the synthetic datasets

| Dataset Name | Samples | Features | Centers | Standard Deviation | Noise | Remarks | No. of clusters (Elbow Method) | No. of clusters (Silhouette Method) |
|---|---|---|---|---|---|---|---|---|
| Aggregated Clusters | 300 | 2 | 8 | 0.3 | - | - | 4 | 6 |
| Anisotropic | 400 | 6 | 3 | [0.5, 1.5, 0.5] | - | - | 3 | 3 |
| Blobs | 100 | 4 | 4 | 1.0 | - | - | 4 | 4 |
| Broken Rings | - | - | - | - | 0.1 | Concentric broken rings | 4 | 10 |
| Checkerboard | 300 | - | - | - | - | 9x9 grid | 4 | 4 |
| Circles | 250 | - | - | - | 0.05 | - | 4 | 10 |
| Concentric Spheres | - | - | - | - | 0.02 | Multiple circles with scaling factors | 4 | 6 |
| Disjoint Clusters | 300 | 2 | 3 | 1.0 | - | Clearly separated clusters | 3 | 3 |
| Feature Entanglement | 300 | 2 | - | - | - | Binary features with XOR pattern | 4 | 4 |
| Friedman's Function | 1000 | - | - | - | - | Complex regression function | 5 | 10 |
| Gaussian Mixture | 500 | 3 | 5 | [0.5, 1.0, 1.5, 2.0, 2.5] | - | - | 4 | 3 |
| Hierarchical Clusters | Varied | 3 | Custom | [0.5, 1, 0.3, 0.8] | - | Variable sample sizes, predefined centers | 4 | 4 |
| High Dimensional Blobs | 300 | 10 | 5 | - | - | - | 5 | 5 |
| Highly Correlated Features | 300 | - | - | - | - | Features with additive noise | 4 | 8 |
| Interlocking Moons | 800 | - | - | - | 0.2 | - | 4 | 2 |
| Manifold Learning Dataset | 300 | - | - | - | 0.1 | Swiss roll for manifold learning | 3 | 10 |
| Moons | 200 | - | - | - | 0.1 | - | 4 | 2 |
| Nested Clusters | 350 | 2 | Custom | [0.5, 0.2, 0.3] | - | Hierarchical centers | 3 | 2 |
| No structure | 300 | 2 | - | - | - | Random scatter | 4 | 4 |
| Non-spherical Gaussian Mixture | 300 | 2 | 3 | [1.0, 2.0, 3.0] | - | Non-spherical shapes due to variances | 3 | 3 |
| Overlapping Circles | 500 | - | - | - | 0.1 | - | 5 | 7 |
| Periodic Patterns | - | - | - | - | - | Repeating linear scales | 4 | 4 |
| Random Uniform Scatter | 1000 | 2 | - | - | - | Scaled by 100 | 4 | 4 |
| Random Walk Clusters | 1000 | - | - | - | - | Cumulative summation of random changes | 4 | 3 |
| Shifting Variance Clusters | 300 | 2 | 3 | Linearly increasing | - | Standard deviations vary linearly | 3 | 3 |
| Simple Blobs | 300 | 2 | 4 | 1.0 | - | - | 4 | 4 |
| Sine Wave Clusters | 200 | - | - | - | 0.1 | Generated with sinusoidal functions | 4 | 5 |
| Sparse High-Dimensional | 200 | 100 | - | - | - | Binary data, mostly sparse | - | 2 |
| Spirals | - | - | - | - | - | Generated with sinusoidal and cosine functions | 4 | 4 |
| Stretched Blobs | 300 | 2 | 3 | 0.5 | - | Center box from 20 to 100 | 3 | 3 |
| Swiss Roll | 300 | - | - | - | 0.1 | 2D projection of 3D Swiss roll | 3 | 10 |
| Varied Density | 100 | 12 | 3 | [1.0, 2.5, 0.5] | - | - | 3 | 3 |
| Winding Function Clusters | 200 | - | - | - | - | Generated with sinusoidal functions | 4 | 9 |

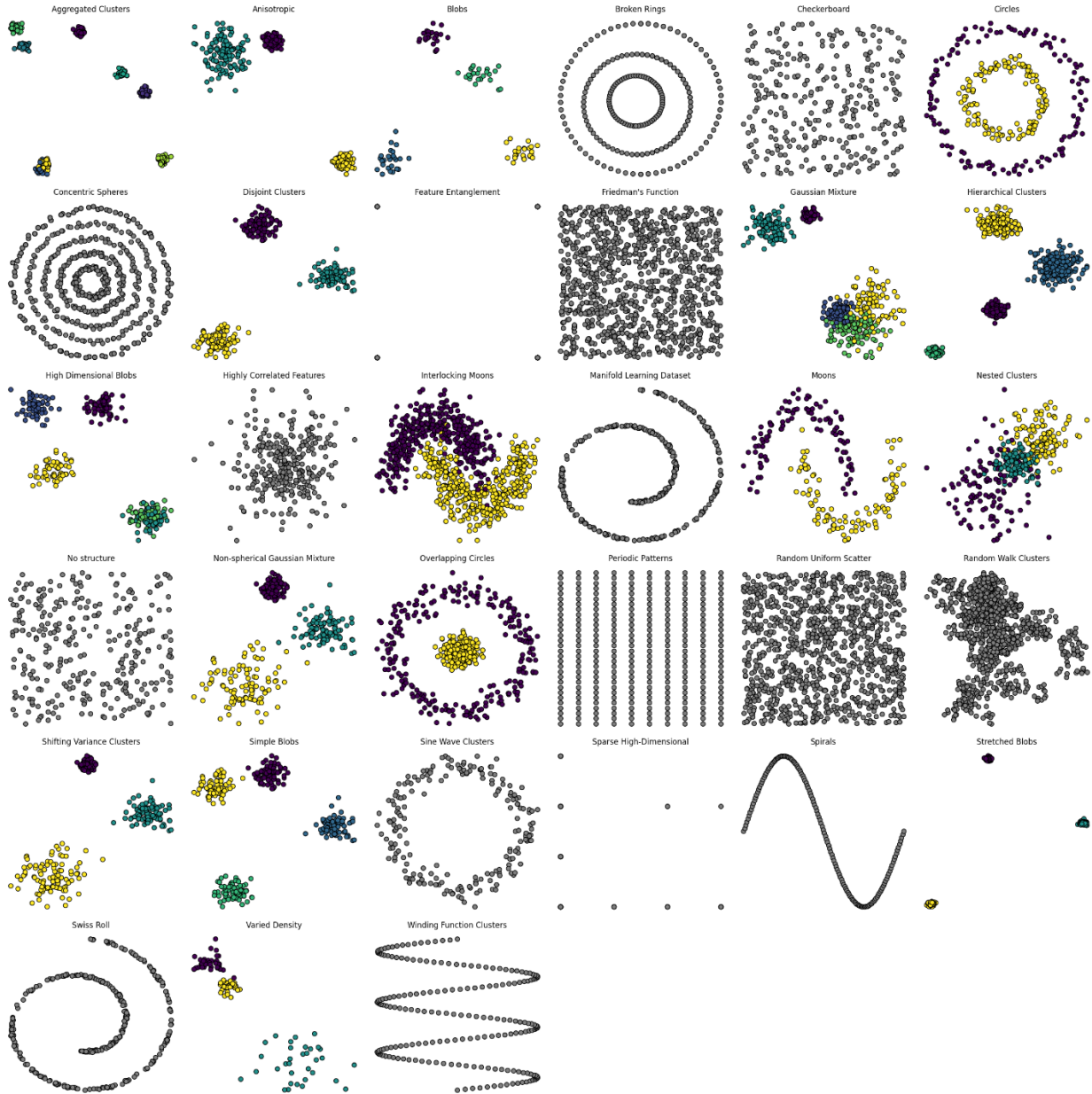

Fig. XXX Visualization of the synthetic datasets [Note: colors denote the datasets with known labels]

The results presented in Table XXX offer a view of the performance of various clustering algorithms across multiple metrics. At the top of the ranking is HDBSCAN, achieving the highest average metric score. This algorithm excels in Homogeneity (0.9099) and Completeness (0.89291), indicating its effectiveness in grouping similar data points together while ensuring that points from the same actual cluster are not split into multiple clusters. In parallel, this algorithm's Silhouette score is also notably high, suggesting that HDBSCAN not only groups data points accurately but does so with clear demarcation between clusters. Following closely are several variants of the SPNIEX algorithm (SPINEX_T, SPINEX_T_PCA, SPINEX_T_TSNE, SPINEX_T_UMAP), all tying for the second through fifth ranks. These variants show a strong consistency in performance across different implementations, whether incorporating PCA, t-SNE, or UMAP for dimensionality reduction. Their scores reflect a well-balanced approach, achieving reasonably high Homogeneity and Completeness metrics, which suggest that these SPNIEX variants can cluster effectively.

K-Means ranks sixth, showcasing its robustness with a high Silhouette score (0.59145), indicating a good separation between clusters. However, it scores slightly lower in Homogeneity and Completeness compared to the SPNIEX variants, which may suggest mixing data points from different true clusters or some data points from the same cluster being spread across multiple clusters. Birch ranks eighth with the highest Homogeneity score, suggesting it is particularly effective at ensuring that clusters contain data points from a single class. However, its lower Completeness score (0.63907) might indicate that it tends to distribute points from the same class into multiple clusters.

In the lower ranks, algorithms like Mean Shift and Agglomerative exhibit varied challenges. Despite a very high Silhouette score (0.74279), Mean Shift has the lowest V-Measure and Homogeneity scores. Similarly, Agglomerative and Ward Hierarchical clustering demonstrate issues with effectively capturing the true data labels, as seen in their lower Homogeneity scores. Interestingly, SPNIEX, in its standard form, ranks the lowest, which may highlight the necessity of its more specialized variants that incorporate additional techniques or optimizations. This could indicate that while the foundational concept of SPNIEX is sound, enhancements and adaptations (like those implemented in its variants) are crucial for optimal performance.

Table XXX Average and overall ranking results on synthetic data

| Algorithm | Silhouette | Calinski-Harabasz | Davies-Bouldin | Homogeneity | Completeness | V-Measure | Mean Across Metrics | Rank |
|---|---|---|---|---|---|---|---|---|
| HDBSCAN | 0.49679 | 0.01314 | 0 | 0.9099 | 0.89291 | 0.89534 | 0.53468 | 1 |
| SPINEX_T | 0.5827 | 0.01314 | 0 | 0.76319 | 0.7884 | 0.75661 | 0.48401 | 2 |
| SPINEX_T_PCA | 0.5827 | 0.01314 | 0 | 0.76319 | 0.7884 | 0.75661 | 0.48401 | 2 |
| SPINEX_T_TSNE | 0.5827 | 0.01314 | 0 | 0.76319 | 0.7884 | 0.75661 | 0.48401 | 2 |
| SPINEX_T_UMAP | 0.5827 | 0.01314 | 0 | 0.76319 | 0.7884 | 0.75661 | 0.48401 | 2 |
| K-Means | 0.59145 | 0.01274 | 0 | 0.80718 | 0.73444 | 0.75671 | 0.48375 | 3 |
| SPINEX_Multi_level | 0.51477 | 0 | 0 | 0.83569 | 0.76365 | 0.77355 | 0.48128 | 4 |
| Birch | 0.57925 | 0.01314 | 0 | 0.93531 | 0.63907 | 0.7179 | 0.48078 | 5 |
| MiniBatch KMeans | 0.58656 | 0.01274 | 0 | 0.79528 | 0.72723 | 0.74813 | 0.47832 | 6 |
| Spectral Clustering | 0.57759 | 0.01274 | 0 | 0.79455 | 0.73459 | 0.74941 | 0.47815 | 7 |
| SPINEX_TSNE | 0.54594 | 0 | 0 | 0.77167 | 0.77086 | 0.75615 | 0.4741 | 8 |
| SPINEX_PCA | 0.54594 | 0 | 0 | 0.77167 | 0.77086 | 0.75615 | 0.4741 | 8 |
| SPINEX_RS | 0.54594 | 0 | 0 | 0.77167 | 0.77086 | 0.75615 | 0.4741 | 8 |
| SPINEX_UMAP | 0.54594 | 0 | 0 | 0.77167 | 0.77086 | 0.75615 | 0.4741 | 8 |
| SPINEX_No.ofClusters | 0.56473 | 0.01314 | 0 | 0.76654 | 0.71506 | 0.73005 | 0.46492 | 9 |
| DBSCAN | 0.55681 | 0.02213 | 0 | 0.73014 | 0.7665 | 0.70578 | 0.46356 | 10 |
| KMedoids | 0.53299 | 0 | 0 | 0.78708 | 0.71305 | 0.7358 | 0.46149 | 11 |
| Affinity Propagation | 0.50205 | 0.03125 | 0 | 0.95538 | 0.5692 | 0.67312 | 0.45517 | 12 |
| Mean Shift | 0.74279 | 0.04204 | 0 | 0.32718 | 0.99708 | 0.37011 | 0.4132 | 13 |
| Agglomerative | 0.51972 | 0 | 0 | 0.43624 | 0.81659 | 0.55884 | 0.38856 | 14 |
| Ward Hierarchical | 0.51972 | 0 | 0 | 0.43624 | 0.81659 | 0.55884 | 0.38856 | 14 |
| OPTICS | 0.38045 | 0.01617 | 0 | 0.67236 | 0.64139 | 0.61624 | 0.38777 | 15 |
| Gaussian Mixture | 0.52613 | 0 | 0 | 0.42691 | 0.80022 | 0.54881 | 0.38368 | 16 |
| SPINEX | 0.45689 | 0 | 0 | 0.45805 | 0.67585 | 0.53377 | 0.35409 | 17 |

Note: SPINEX_T: with PCA as True. SPINEX_T_ TSNE: with PCA as True and approximation method=TSNE, SPINEX_T_UMAP: with PCA as True and approximation method= UMAP, SPINEX_T_ PCA: with PCA as True and approximation method= PCA, SPINEX_RS with approximation

method=random_sampling and sample size=0.5, SPINEX_PCA with approximation method= PCA and sample size=0.5, SPINEX_No.ofClusters with no. of clusters fixed at 4.

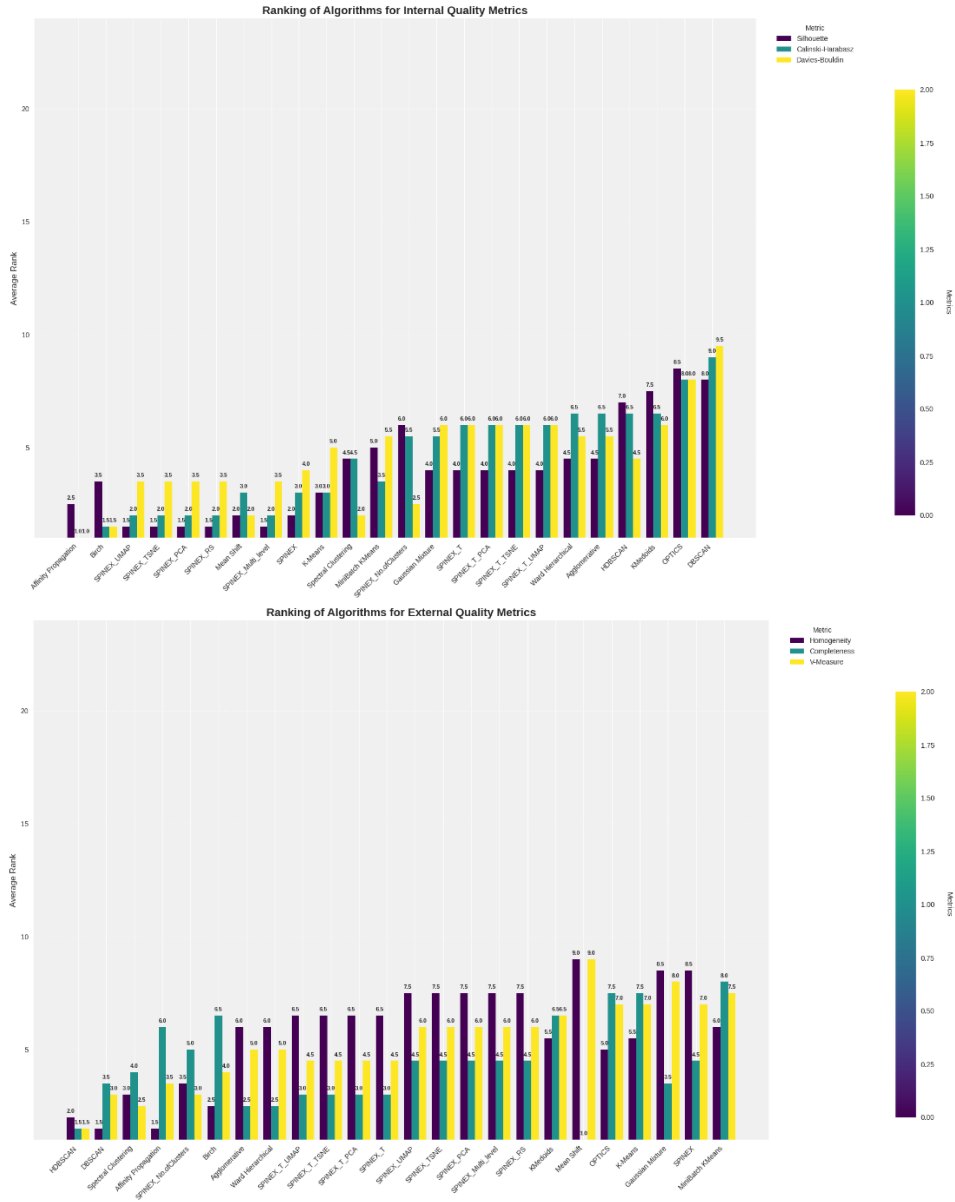

Figure XXX Individual rankings per algorithm for the internal and external metrics.

A Pareto analysis is conducted on the synthetic datasets to identify the best-performing algorithms based on multiple evaluation metrics and execution time. This analysis uses the concept of Pareto optimality to find non-dominated solutions that represent the best trade-offs between different performance metrics: Silhouette, Calinski-Harabasz, Davies-Bouldin, Homogeneity, Completeness, and V-Measure. First, all metrics are then normalized to a 0-1 scale using min-max normalization. After normalization, the normalized data is grouped by both algorithm and dataset to calculate mean values for each metric. These grouped results are then further aggregated by the algorithm to get overall performance across all datasets. A solution is Pareto optimal if no other solution is better in all metrics simultaneously.

The Pareto optimality concept ensures that the final set of recommended algorithms represents truly superior options, each offering a unique balance of strengths across different performance criteria. The final output of this analysis is a set of Pareto optimal solutions, where each algorithm in this set offers a unique trade-off between the different performance criteria. In practical terms, this analysis helps in selecting clustering algorithms that offer the best balance of performance across multiple criteria.

Table XXX Pareto analysis of synthetic data

| Algorithm | Silhouette | Calinski-Harabasz | Davies-Bouldin | Homogeneity | Completeness | V-Measure |
|---|---|---|---|---|---|---|
| Affinity Propagation | 0.50205 | 3.13E-02 | 0 | 0.95538 | 0.5692 | 0.67312 |
| Agglomerative | 0.51972 | 7.44E-31 | 0 | 0.43624 | 0.81659 | 0.55884 |
| Birch | 0.57925 | 1.31E-02 | 0 | 0.93531 | 0.63907 | 0.7179 |
| DBSCAN | 0.55681 | 2.21E-02 | 0 | 0.73014 | 0.7665 | 0.70578 |
| Gaussian Mixture | 0.52613 | 7.70E-31 | 0 | 0.42691 | 0.80022 | 0.54881 |
| HDBSCAN | 0.49679 | 1.31E-02 | 0 | 0.9099 | 0.89291 | 0.89534 |
| K-Means | 0.59145 | 1.27E-02 | 0 | 0.80718 | 0.73444 | 0.75671 |
| KMedoids | 0.53299 | 1.62E-29 | 0 | 0.78708 | 0.71305 | 0.7358 |
| Mean Shift | 0.74279 | 4.20E-02 | 0 | 0.32718 | 0.99708 | 0.37011 |
| SPINEX_Multi_level | 0.51477 | 2.16E-29 | 0 | 0.83569 | 0.76365 | 0.77355 |
| SPINEX_No.ofClusters | 0.56473 | 1.31E-02 | 0 | 0.76654 | 0.71506 | 0.73005 |
| SPINEX_PCA | 0.54594 | 2.22E-29 | 0 | 0.77167 | 0.77086 | 0.75615 |
| SPINEX_T_UMAP | 0.5827 | 1.31E-02 | 0 | 0.76319 | 0.7884 | 0.75661 |
| Spectral Clustering | 0.57759 | 1.27E-02 | 0 | 0.79455 | 0.73459 | 0.74941 |
| Affinity Propagation | 0.50205 | 3.13E-02 | 0 | 0.95538 | 0.5692 | 0.67312 |
| Agglomerative | 0.51972 | 7.44E-31 | 0 | 0.43624 | 0.81659 | 0.55884 |
| Birch | 0.57925 | 1.31E-02 | 0 | 0.93531 | 0.63907 | 0.7179 |
| DBSCAN | 0.55681 | 2.21E-02 | 0 | 0.73014 | 0.7665 | 0.70578 |

Figure XXX presents a visual example of the clusters predicted by SPINEX and other algorithms on the Moons dataset. Please note that plots for other datasets are shown in the Appendix.

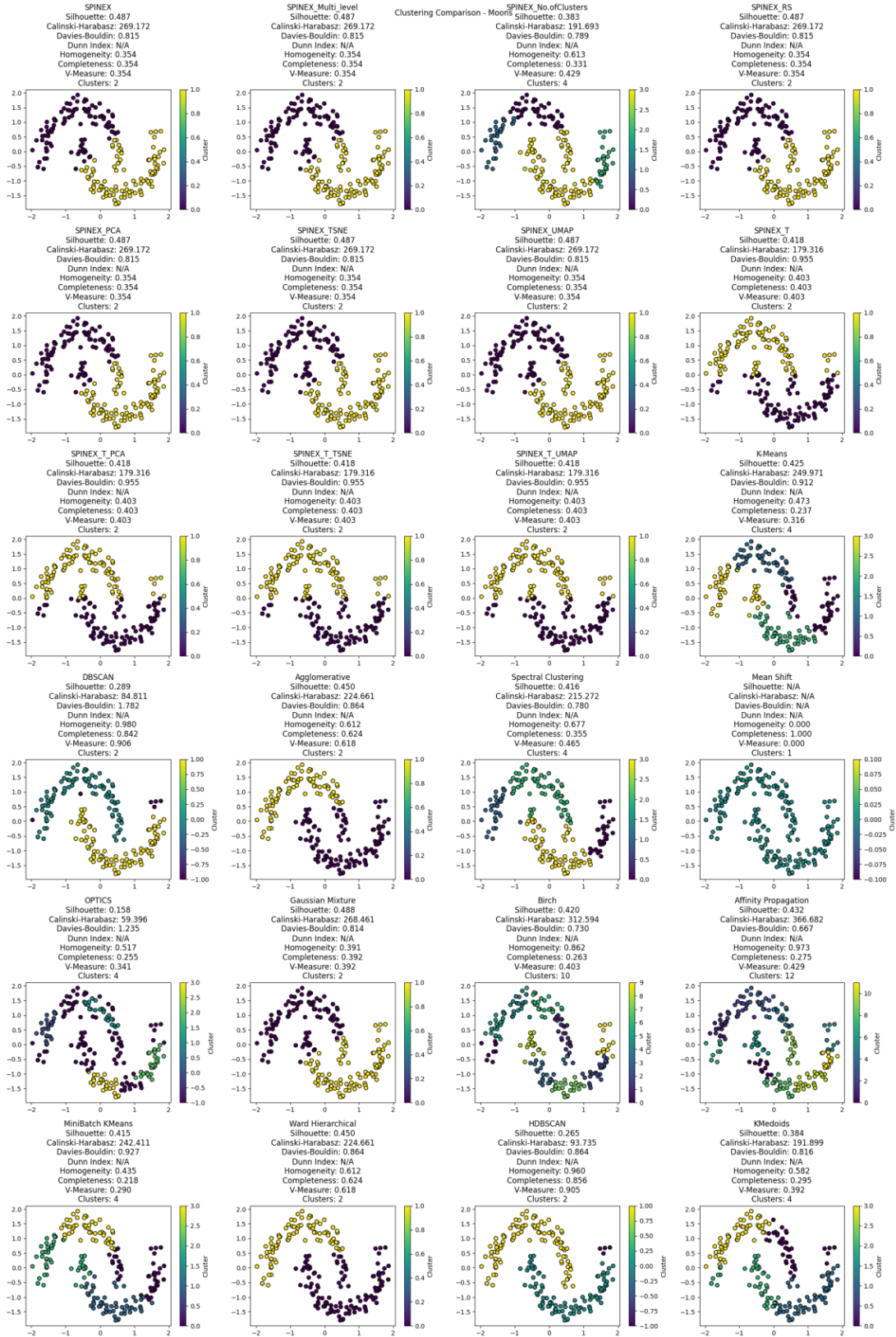

Fig. 4 Visualization of clusters predicted across the Moons dataset

*4.2 Real datasets*

Eighteen real datasets were used in this leg of the analysis (which followed the same procedure used in the former section). The selected datasets comprise numerous problems and scenarios (see Table XXX and Fig. XXX), with additional details that can be found in the original sources within the Scikit [40] and Seaborn [41] libraries.

Table XXX Real datasets used in the analysis

| Dataset Name | Samples | Features | References |
|---|---|---|---|
| Wine | 178 | 13 | [40] |
| Olivetti Faces | 400 | 4096 | [40] |
| Diabetes | 442 | 10 | [40] |
| Breast Cancer | 569 | 30 | [40] |
| Seaborn Car Crashes | 51 | 8 | [41] |
| Seaborn Diamonds | 53,940 | 7 | [41] |
| Seaborn Dots | 308 | 4 | [41] |
| Seaborn Exercise | 90 | 6 | [41] |
| Seaborn Flights | 144 | 3 | [41] |
| Seaborn Fmri | 1,068 | 5 | [41] |
| Seaborn Iris | 150 | 4 | [41] |
| Seaborn Mpg | 398 | 7 | [41] |
| Seaborn Penguins | 344 | 5 | [41] |
| Seaborn Planets | 1,035 | 6 | [41] |
| Seaborn Tips | 244 | 7 | [41] |
| Seaborn Titanic | 891 | 6 | [41] |
| Seaborn Anscombe | 44 | 2 | [41] |
| Seaborn Attention | 60 | 5 | [41] |

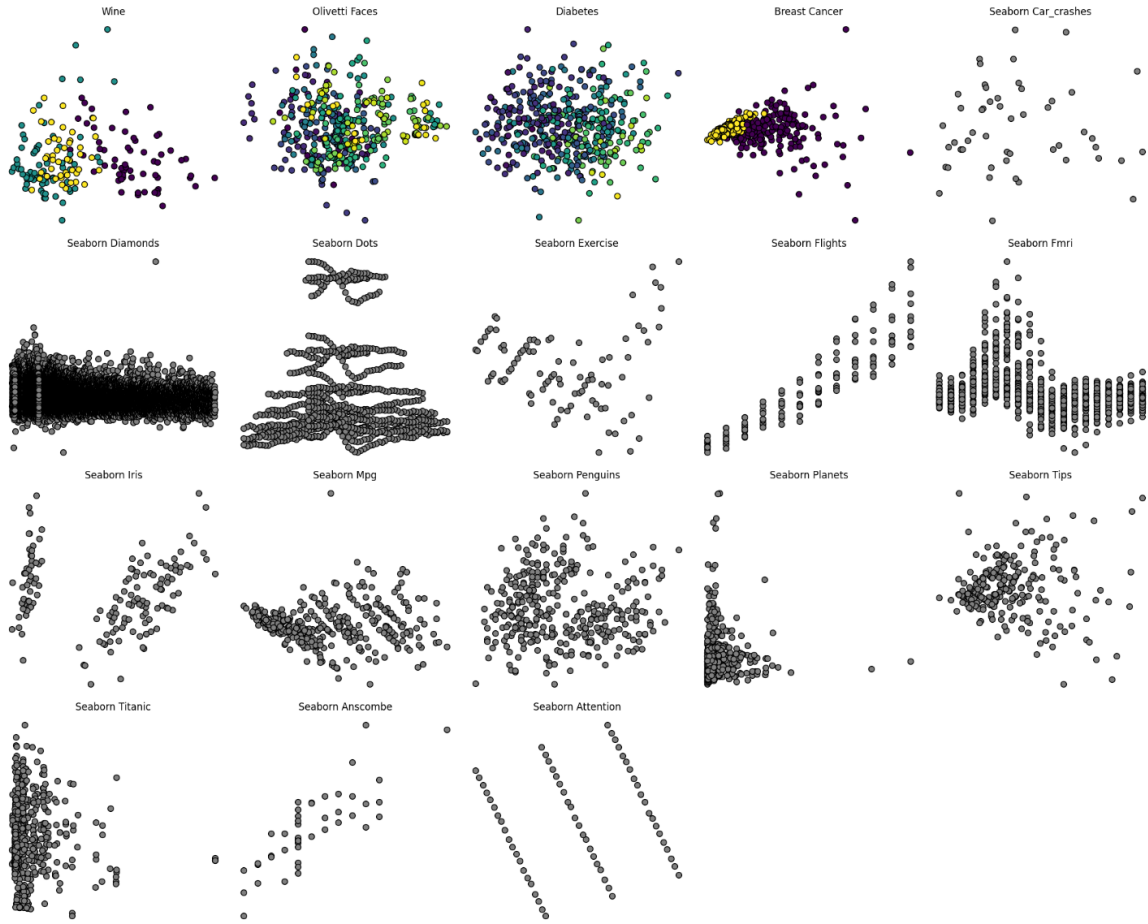

Fig. XXX Visualization of the real datasets

The results of the benchmarking and ranking analysis are listed in Table XXX. This table shows the performance of various SPINEX versions against the selected clustering algorithms. Overall, Birch, Mean Shift, and Affinity Propagation rank as the top 3 algorithms, followed closely by SPINEX. It is worth noting that six variants from SPINEX reside in the top-10 performing algorithms. The best performing SPNIEX variant is the SPINEX_T, which accommodates dimensionality reduction by using PCA. Furthermore, the default variant of SPINEX managed to perform better than HDBSCAN, OPTICS, and DBSCAN. Figure XXX further shows the individual rankings per algorithm for the internal and external metrics.

Table XXX Average and overall ranking results on real data

| Algorithm | Silhou-ette | Calinski-Harabasz | Davies-Bouldin | Homogeneity | Completeness | V-Measure | Mean Across Metrics | Rank |
|---|---|---|---|---|---|---|---|---|
| Birch | 0.6651 | 0.1604 | 0.3367 | 0.9969 | 0.4288 | 0.6034 | 0.5319 | 1 |
| Mean Shift | 0.6898 | 0.1014 | 0.2347 | 0.8324 | 0.4366 | 0.5438 | 0.4731 | 2 |
| Affinity Propagation | 0.6623 | 0.1225 | 0.1843 | 0.7568 | 0.5075 | 0.6012 | 0.4724 | 3 |
| SPINEX_T | 0.6027 | 0.1048 | 0.1783 | 0.6503 | 0.6396 | 0.6570 | 0.4721 | 4 |
| SPINEX_T_UMAP | 0.6019 | 0.1047 | 0.1775 | 0.6449 | 0.6422 | 0.6554 | 0.4711 | 4 |
| SPINEX_T_TSNE | 0.6024 | 0.1048 | 0.1781 | 0.6481 | 0.6382 | 0.6550 | 0.4711 | 4 |
| SPINEX_T_PCA | 0.6022 | 0.1048 | 0.1783 | 0.6452 | 0.6367 | 0.6527 | 0.4700 | 5 |
| SPINEX_Multi_level | 0.5733 | 0.1040 | 0.2314 | 0.6967 | 0.5295 | 0.6018 | 0.4561 | 6 |
| K-Means | 0.7213 | 0.1946 | 0.1715 | 0.4828 | 0.5666 | 0.5095 | 0.4411 | 7 |

| Algorithm | | | | | | | Rank |
|---|---|---|---|---|---|---|---|
| SPINEX_TSNE | 0.6173 | 0.1249 | 0.1968 | 0.5171 | 0.5875 | 0.5386 | 0.4304 | 8 |
| SPINEX_UMAP | 0.6164 | 0.1250 | 0.1964 | 0.5164 | 0.5879 | 0.5380 | 0.4300 | 9 |
| SPINEX_PCA | 0.6171 | 0.1249 | 0.1967 | 0.5160 | 0.5856 | 0.5371 | 0.4296 | 10 |
| Gaussian Mixture | 0.7508 | 0.1686 | 0.1736 | 0.3334 | 0.7036 | 0.4399 | 0.4283 | 11 |
| SPINEX_RS | 0.6144 | 0.1249 | 0.1970 | 0.5121 | 0.5818 | 0.5324 | 0.4271 | 12 |
| MiniBatch KMeans | 0.7046 | 0.1813 | 0.1629 | 0.4461 | 0.5450 | 0.4764 | 0.4194 | 13 |
| SPINEX_No.ofClusters | 0.6737 | 0.1510 | 0.1591 | 0.4443 | 0.5585 | 0.4743 | 0.4101 | 14 |
| Agglomerative | 0.7675 | 0.1760 | 0.1845 | 0.2736 | 0.6433 | 0.3749 | 0.4033 | 15 |
| Ward Hierarchical | 0.7675 | 0.1760 | 0.1845 | 0.2736 | 0.6433 | 0.3749 | 0.4033 | 15 |
| KMedoids | 0.6673 | 0.1383 | 0.1517 | 0.4172 | 0.4779 | 0.4252 | 0.3796 | 16 |
| Spectral Clustering | 0.7666 | 0.1039 | 0.2878 | 0.1509 | 0.5810 | 0.2124 | 0.3504 | 17 |
| SPINEX | 0.5033 | 0.0956 | 0.0937 | 0.3702 | 0.4807 | 0.4143 | 0.3263 | 18 |
| HDBSCAN | 0.5268 | 0.0461 | 0.1092 | 0.2387 | 0.4493 | 0.3155 | 0.2809 | 19 |
| OPTICS | 0.5383 | 0.0560 | 0.0917 | 0.1007 | 0.6964 | 0.1471 | 0.2717 | 20 |
| DBSCAN | 0.2941 | 0.0160 | 0.0966 | 0.0000 | 1.0000 | 0.0000 | 0.2345 | 21 |

Note: SPINEX_T: with PCA as True. SPINEX_T_ TSNE: with PCA as True and approximation method=TSNE, SPINEX_T_UMAP: with PCA as True and approximation method= UMAP, SPINEX_T_ PCA: with PCA as True and approximation method= PCA, SPINEX_RS with approximation method=random_sampling and sample size=0.5, SPINEX_PCA with approximation method= PCA and sample size=0.5, SPINEX_No.ofClusters with no. of clusters fixed at 4.

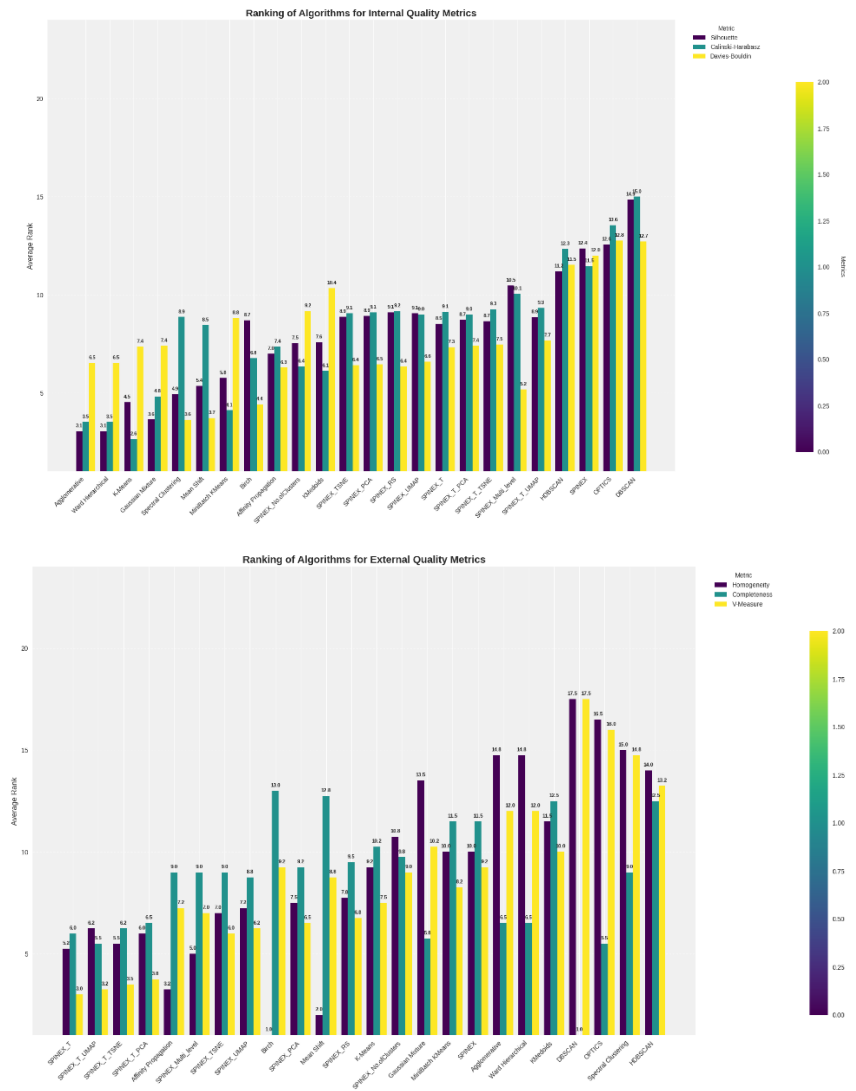

Figure XXX Individual rankings per algorithm for the internal and external metrics.

Similar to the Pareto analysis conducted in the synthetic data, a companion analysis is also conducted. This analysis aims to identify the best-performing algorithms based on multiple evaluation metrics and execution time. The outcome of this analysis is listed in Table XXX and identifies nine SPINEX variants as Pareto optimal.

Table XXX Pareto analysis of real data

| Algorithm | Silhouette | Calinski-Harabasz | Davies-Bouldin | Homogeneity | Completeness | V-Measure |
|---|---|---|---|---|---|---|
| Affinity Propagation | 0.6623 | 0.1225 | 0.1843 | 0.7568 | 0.5075 | 0.6012 |
| Agglomerative | 0.7675 | 0.1760 | 0.1845 | 0.2736 | 0.6433 | 0.3749 |
| Birch | 0.6651 | 0.1604 | 0.3367 | 0.9969 | 0.4288 | 0.6034 |
| DBSCAN | 0.2941 | 0.0160 | 0.0966 | 0.0000 | 1.0000 | 0.0000 |
| Gaussian Mixture | 0.7508 | 0.1686 | 0.1736 | 0.3334 | 0.7036 | 0.4399 |
| K-Means | 0.7213 | 0.1946 | 0.1715 | 0.4828 | 0.5666 | 0.5095 |
| KMedoids | 0.6673 | 0.1383 | 0.1517 | 0.4172 | 0.4779 | 0.4252 |
| Mean Shift | 0.6898 | 0.1014 | 0.2347 | 0.8324 | 0.4366 | 0.5438 |
| SPINEX_Multi_level | 0.5733 | 0.1040 | 0.2314 | 0.6967 | 0.5295 | 0.6018 |
| SPINEX_No.ofClusters | 0.6737 | 0.1510 | 0.1591 | 0.4443 | 0.5585 | 0.4743 |
| SPINEX_RS | 0.6144 | 0.1249 | 0.1970 | 0.5121 | 0.5818 | 0.5324 |
| SPINEX_T | 0.6027 | 0.1048 | 0.1783 | 0.6503 | 0.6396 | 0.6570 |
| SPINEX_TSNE | 0.6173 | 0.1249 | 0.1968 | 0.5171 | 0.5875 | 0.5386 |
| SPINEX_T_PCA | 0.6022 | 0.1048 | 0.1783 | 0.6452 | 0.6367 | 0.6527 |
| SPINEX_T_TSNE | 0.6024 | 0.1048 | 0.1781 | 0.6481 | 0.6382 | 0.6550 |
| SPINEX_T_UMAP | 0.6019 | 0.1047 | 0.1775 | 0.6449 | 0.6422 | 0.6554 |
| SPINEX_UMAP | 0.6164 | 0.1250 | 0.1964 | 0.5164 | 0.5879 | 0.5380 |
| Spectral Clustering | 0.7666 | 0.1039 | 0.2878 | 0.1509 | 0.5810 | 0.2124 |

Figure XXX presents a visual example of the clusters predicted by SPINEX and other algorithms on the Seaborn Flights dataset. Please note that the plots for all datasets are shown in the Appendix.

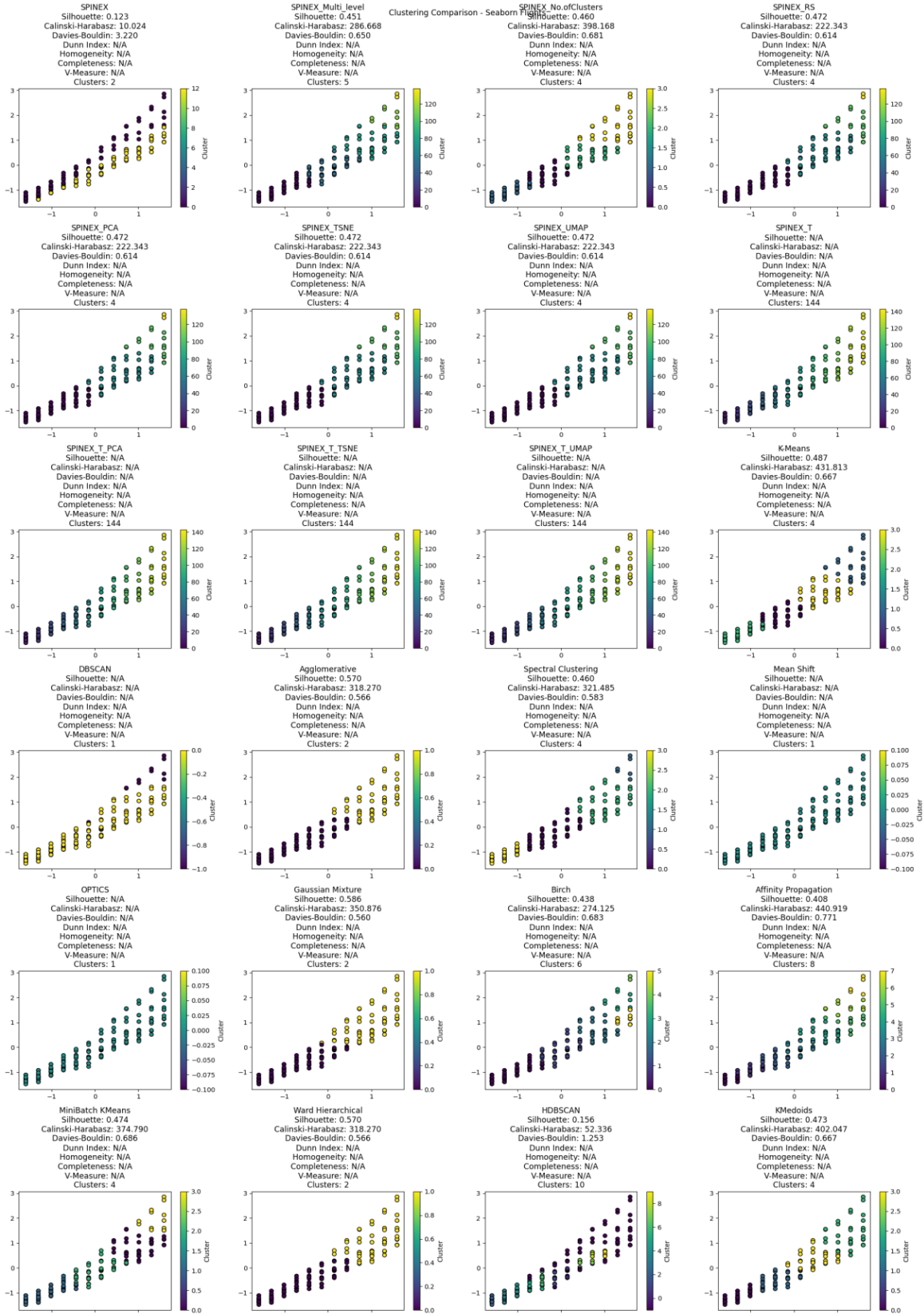

Fig. XXX Visualization of clusters predicted on the Seaborn Flights dataset

*4.3 Complexity analysis*

Here, we conduct a comparative complexity analysis. This analysis was carried out across 100, 1000, and 10000 samples (n) with 50, 100, 500, and 1000 features (d). The main components of the complexity analysis are encapsulated in the following functions:

- The measure_execution_time initializes each algorithm and records its running time across 30 trials. Execution times are captured, ensuring a minimum reportable time to avoid logging errors due to extremely fast executions.
- The estimate_complexity function then uses these timing results to estimate the algorithmic complexity. Input data sizes and corresponding times are logarithmically transformed to linearize potential exponential relationships, facilitating linear regression analysis through curve_fit. The function classifies the complexity based on the slope of the fitted line, categorizing it into common complexity classes such as O(1), O(log n), O(n), O(n log n), and so forth, depending on the slope's value.
- The run_complexity_analysis function examines the complexity of each algorithm across various configurations of feature and sample sizes is systematically evaluated. This function iterates over multiple algorithms and configurations, invoking measure_execution_time for each setting and accumulating the results.

This analysis shows that SPINEX, with an empirical complexity in the range of $O(n^{0.22} \times d) - O(n^{1.46} \times d)$. This complexity matches well with some of the selected algorithms, as can be seen in Figure XXX.

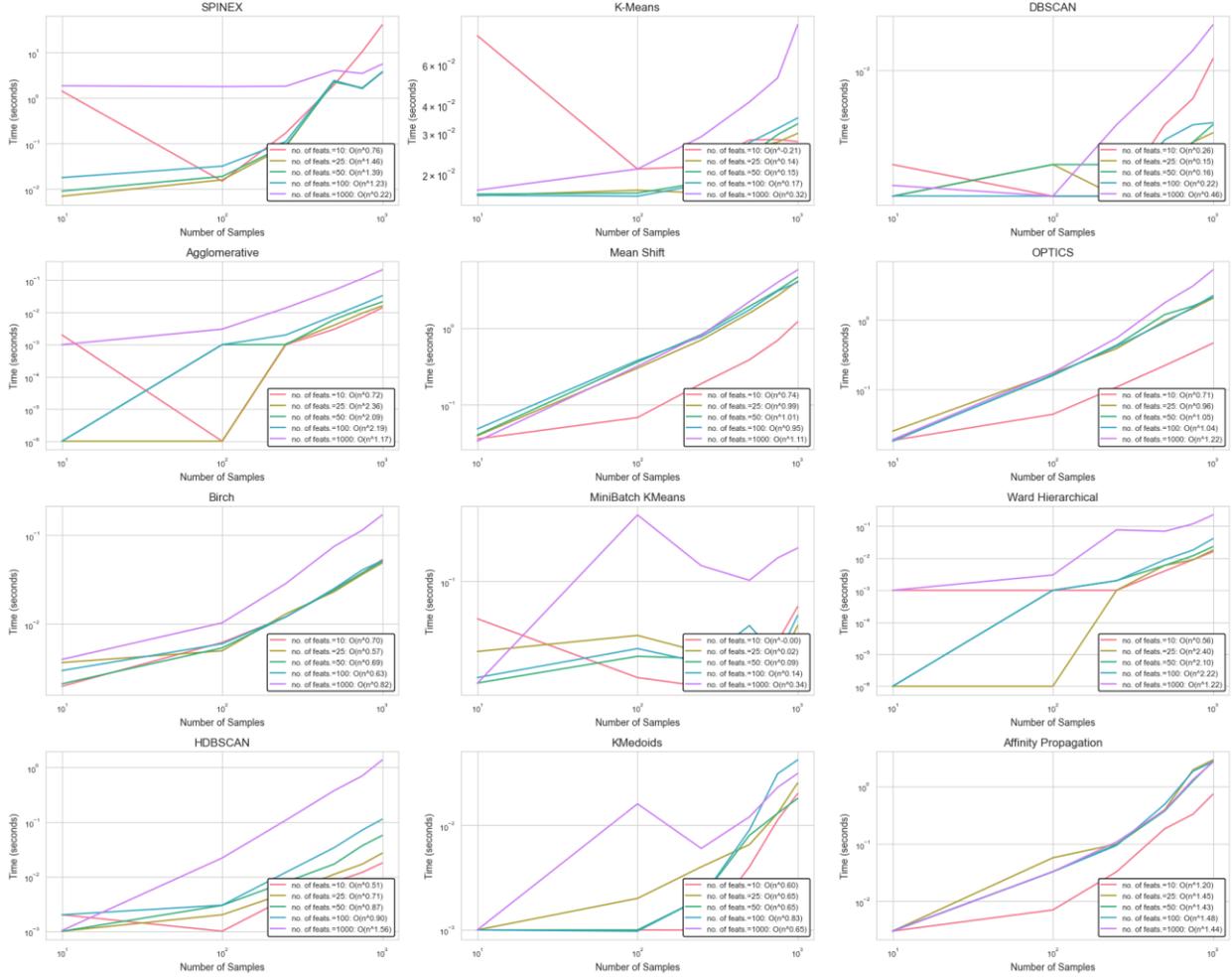

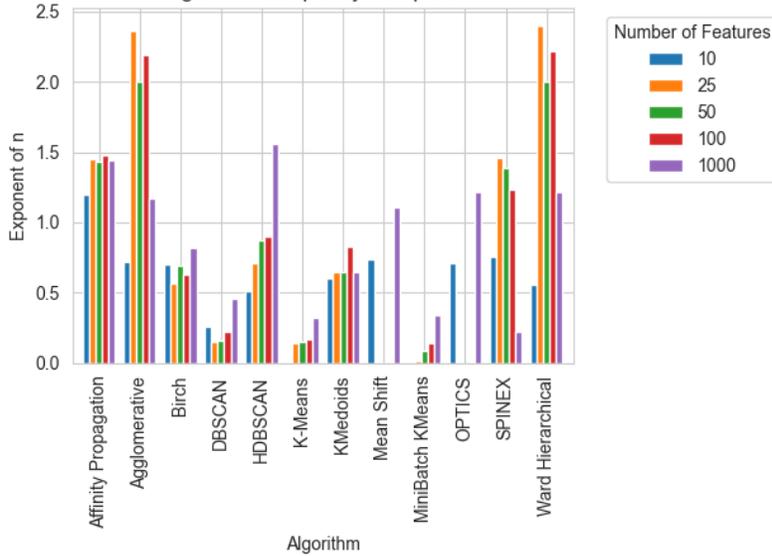

Fig. XXX Complexity analysis

*4.4 Explainability analysis*

To showcase the explainability capabilities of SPINEX as a clustering analysis, an example of noting the clustering of observation 0 in the Moons dataset is shown herein. Figure XXX shows the selected neighbors for this observation when all similarity methods are used (top) and when the cosine method (bottom) is used. As one can see, the neighbors update according to the selected similarity method. The same figure shows the contribution of each of the two features to the selected cluster.

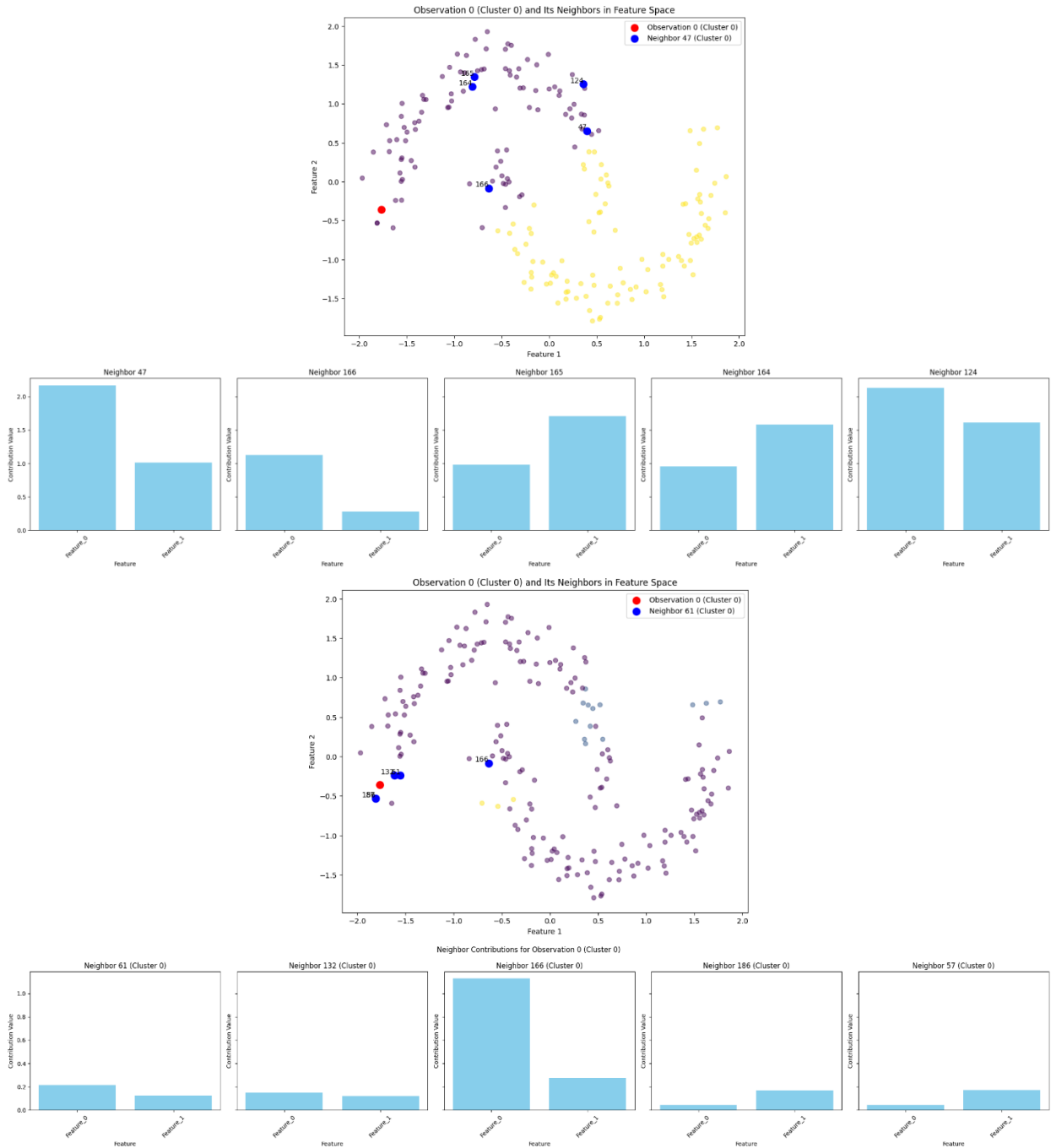

Fig. XXX Example of explainability

## 5.0 Future research directions

Despite its diversity and sophistication, the current landscape of clustering algorithms presents several inherent limitations that pave the way for future research and development. For example, future research in clustering algorithms should focus on addressing the limitations of current methods while leveraging their strengths. For example, for partition-based algorithms like K-Means and K-Medoids, efforts should be directed toward developing more robust initialization techniques, adaptive methods for determining optimal cluster numbers, and integration with kernel methods to handle non-linear cluster boundaries [42]. Similarly, for density-based methods such as DBSCAN, OPTICS, and HDBSCAN, research should concentrate on adaptive parameter selection techniques that respond to data density and distribution characteristics [43].

Hierarchical and spectral clustering methods offer unique insights into data structure but face scalability issues [44]. Future work can explore hybrid models that combine hierarchical approaches with density or grid-based techniques to reduce computational demands while preserving interpretative benefits. Developing scalable algorithms capable of handling large similarity matrices and methods for automatic, robust graph construction are essential for spectral clustering. Gaussian Mixture Models could benefit from more robust model selection criteria and extensions to non-Gaussian mixtures, broadening their applicability across various data distributions.

Exploring incremental or online versions of clustering algorithms to adapt to streaming data represents another promising avenue. Furthermore, developing more sophisticated mini-batch sampling techniques for algorithms like MiniBatch K-Means could improve efficiency without sacrificing clustering quality. Lastly, as datasets continue to grow in dimensionality, size, and complexity, research into parallel and distributed implementations of clustering algorithms will become increasingly important. This may involve leveraging advances in hardware acceleration and cloud computing to enable clustering at multi-scales [45].

## 6.0 Conclusions

This paper introduces a novel member of the SPINEX (Similarity-based Predictions with Explainable Neighbors Exploration) family, which enhances clustering performance by leveraging the concept of similarity and higher-order interactions across multiple subspaces. The effectiveness of the proposed SPINEX variant was rigorously evaluated through a comprehensive benchmarking study involving 13 well-established clustering algorithms. Our extensive testing encompassed 51 diverse datasets, spanning various domains, dimensions, and complexities, to assess SPINEX's capabilities robustly. The findings from our experiments indicate that SPINEX consistently ranks within the top-5 best-performing algorithms, showcasing its efficacy in clustering analysis while maintaining moderate computational complexity on the order between $O(n^{0.22} \times d) - O(n^{1.46} \times d)$. Furthermore, this study also highlights the explainability features of SPINEX, offering insights into the clustering process by exploring explainable neighbors. This aspect of SPINEX facilitates a deeper understanding of the model's decisions and enhances its applicability in scenarios where interpretability is crucial.

**Data Availability**

Some or all data, models, or code that support the findings of this study are available from the corresponding author upon reasonable request.

SPINEX can be accessed from [**to be added**].

**Conflict of Interest**

The authors declare no conflict of interest.

**Appendix**

Python script, additional results, and visualizations.

A. Python script: to be provided.

184 B. Sample visualization of synthetic datasets

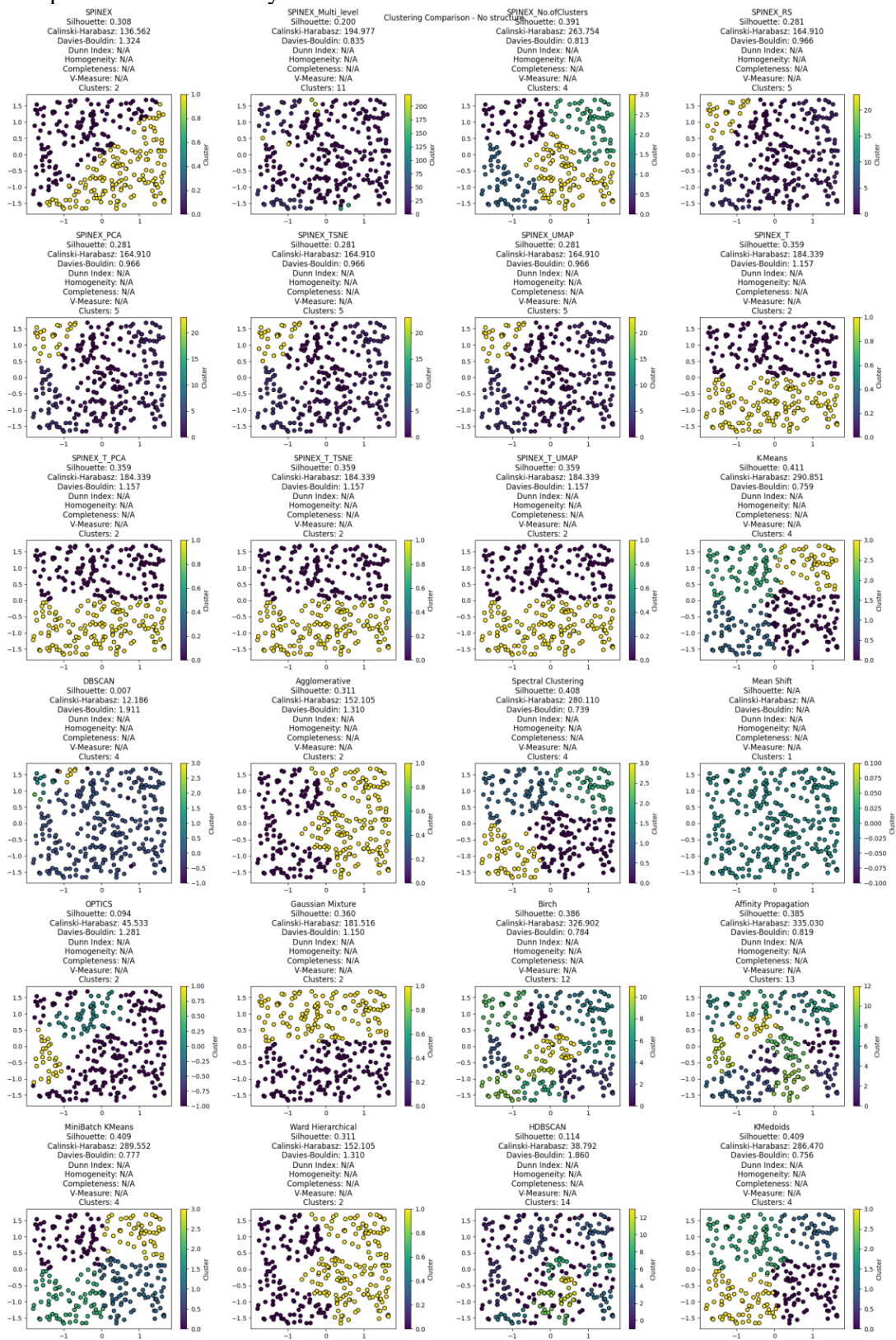

185 C.

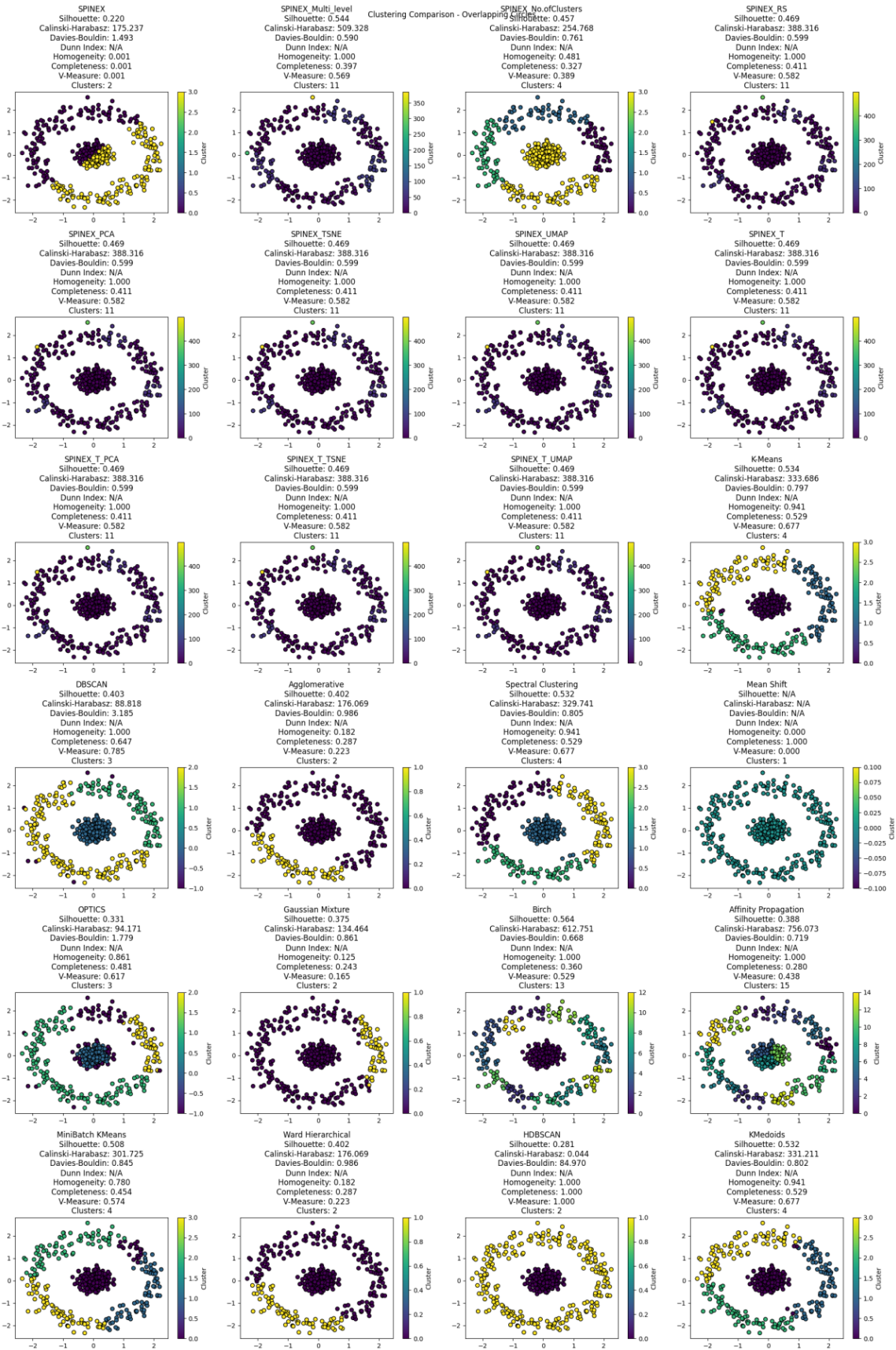

186

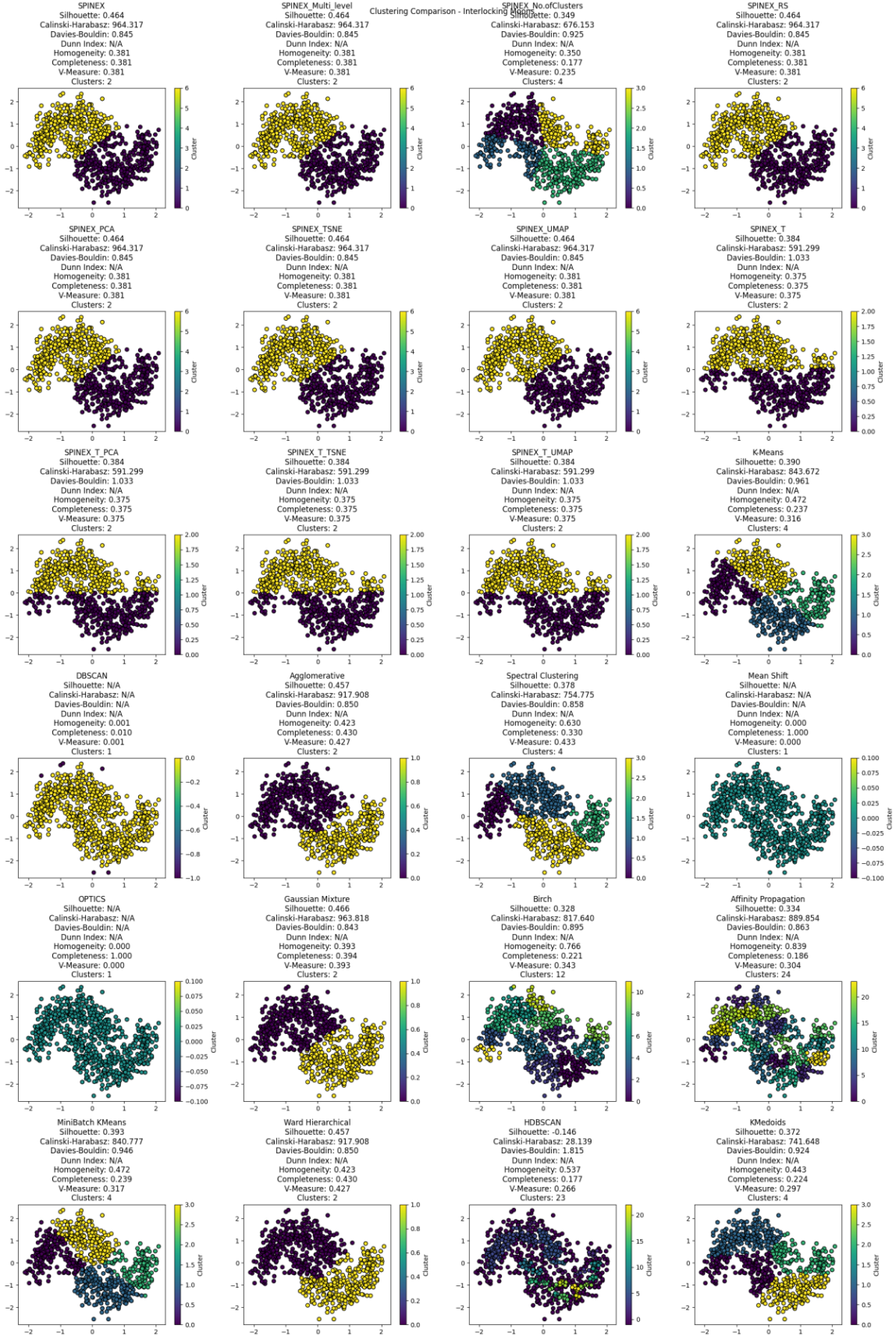

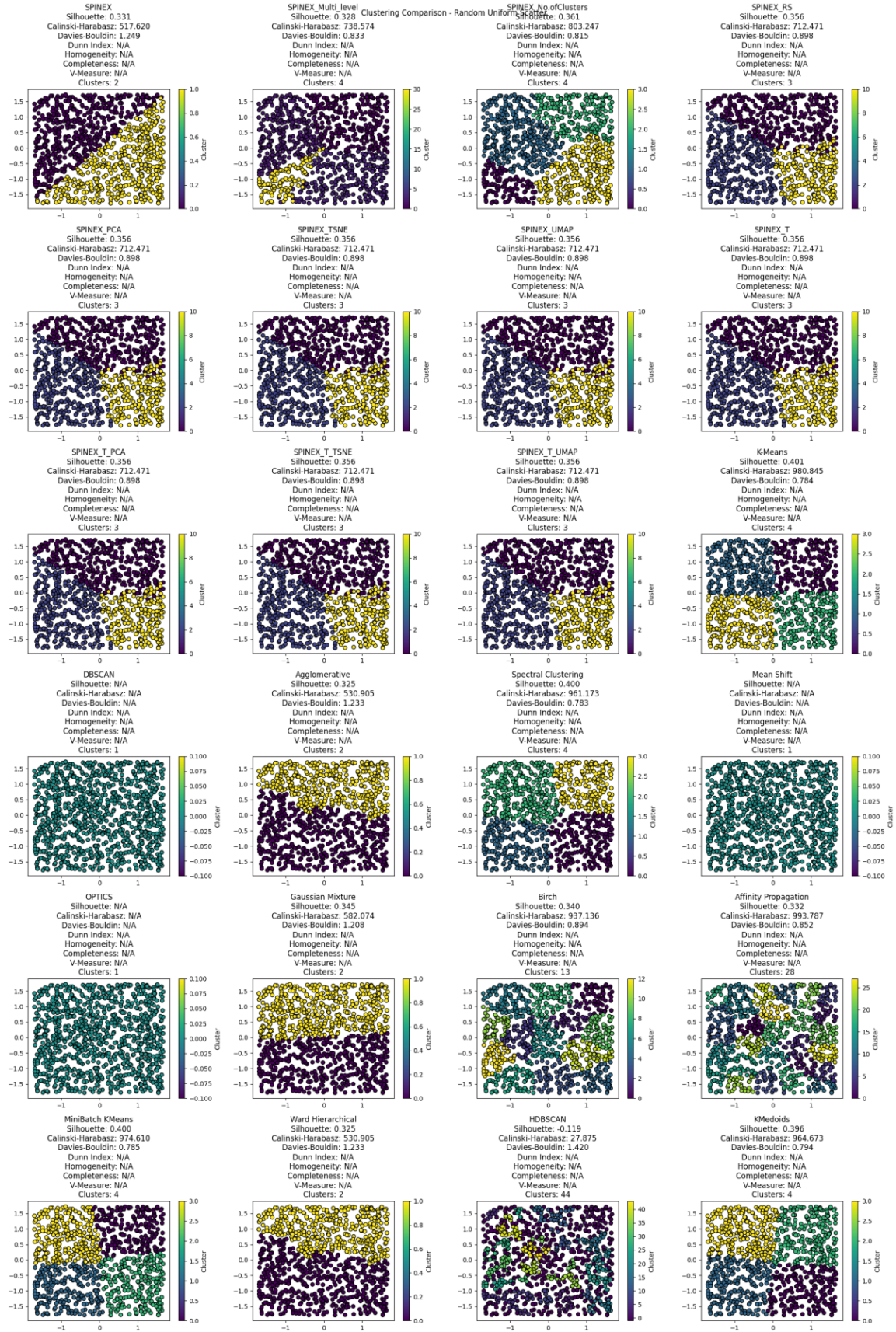
188

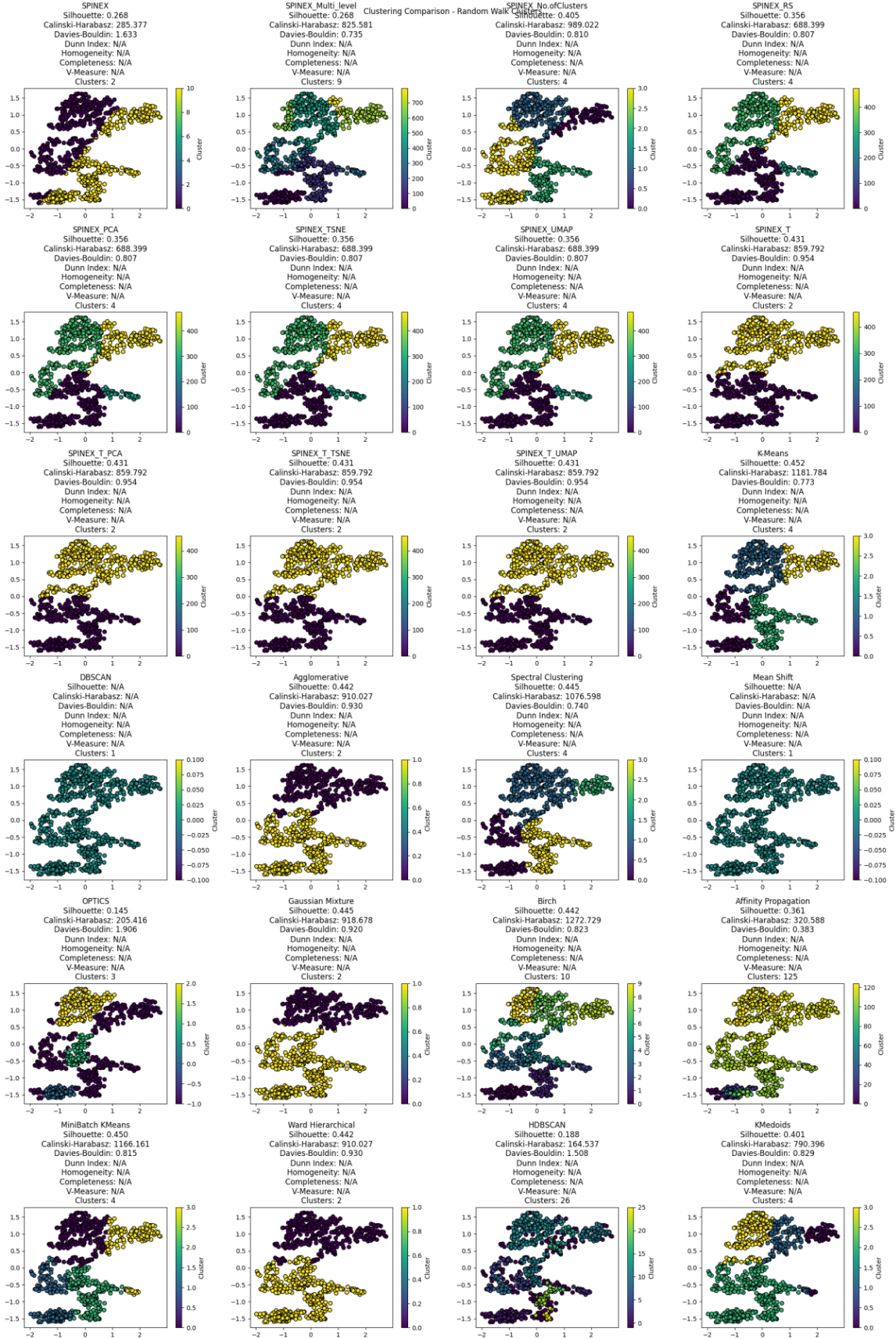

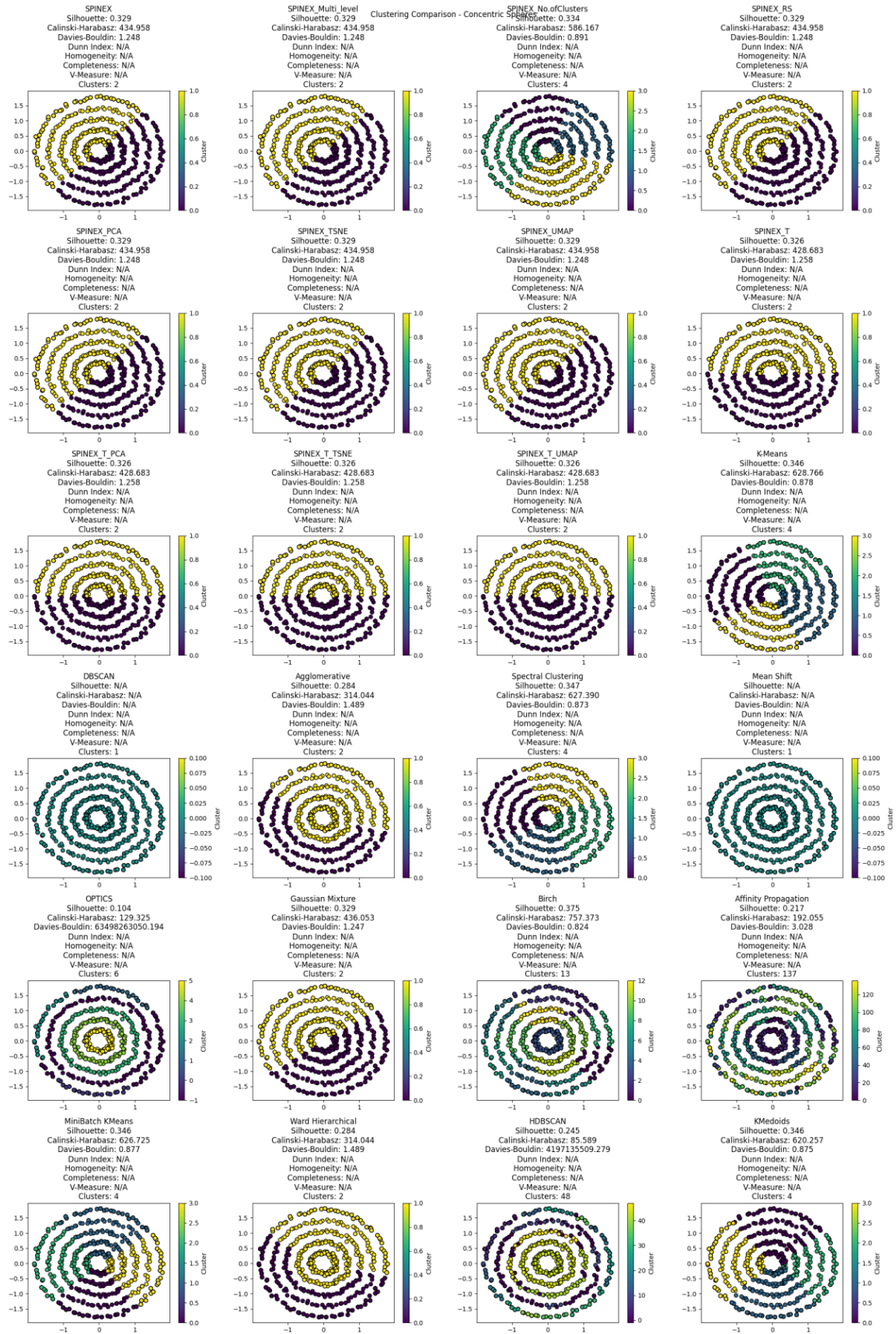

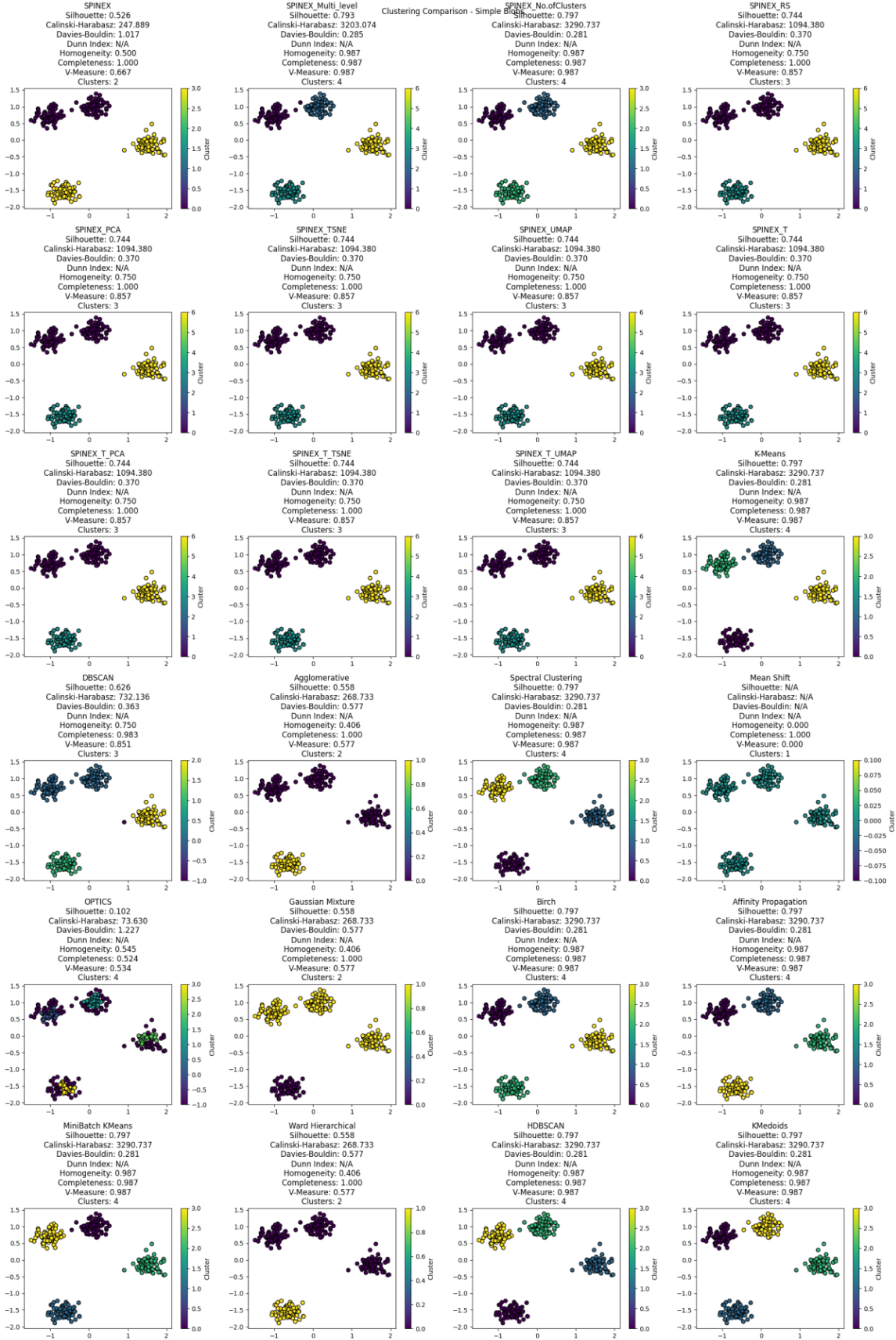

192    C. Sample visualization of real datasets

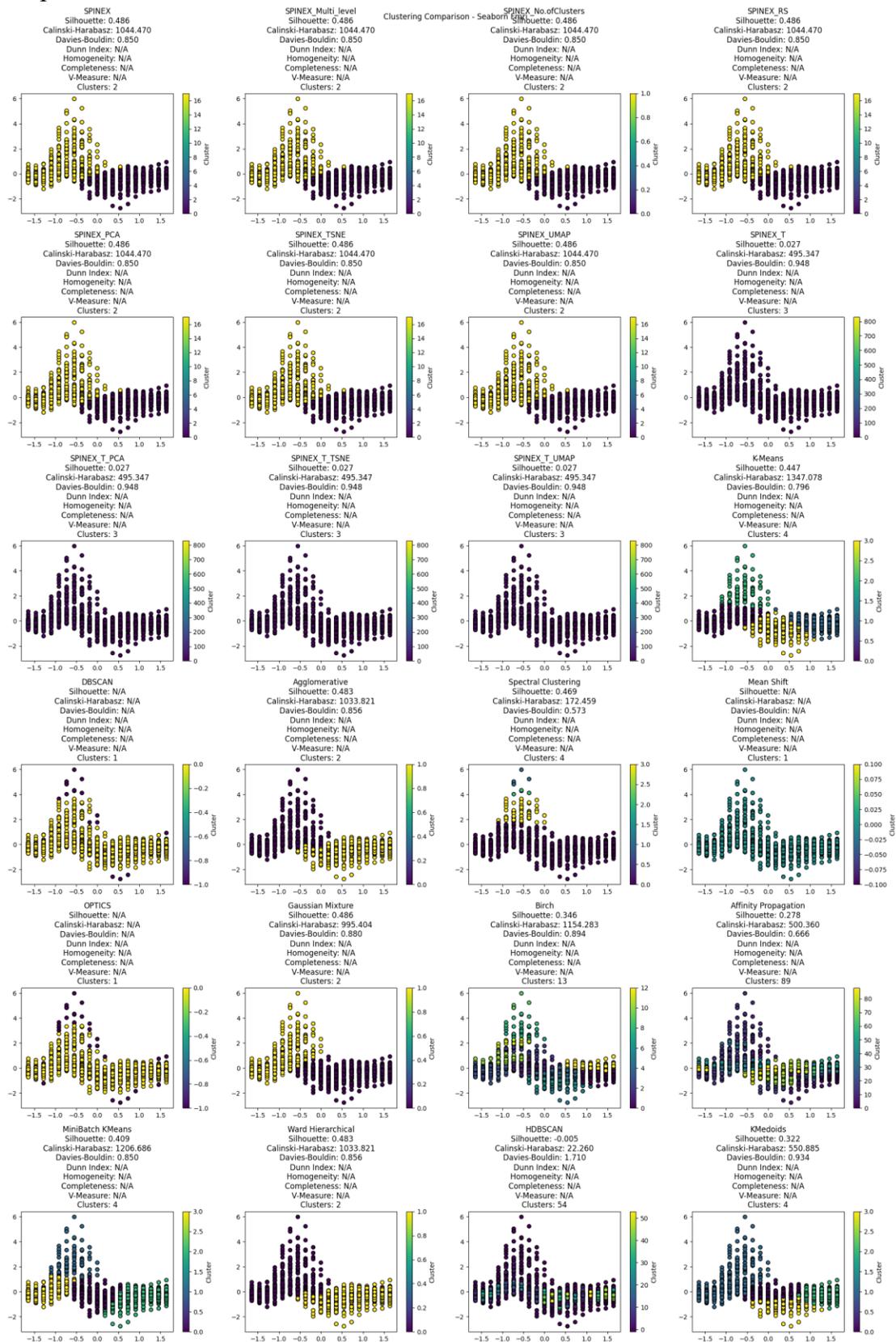

193

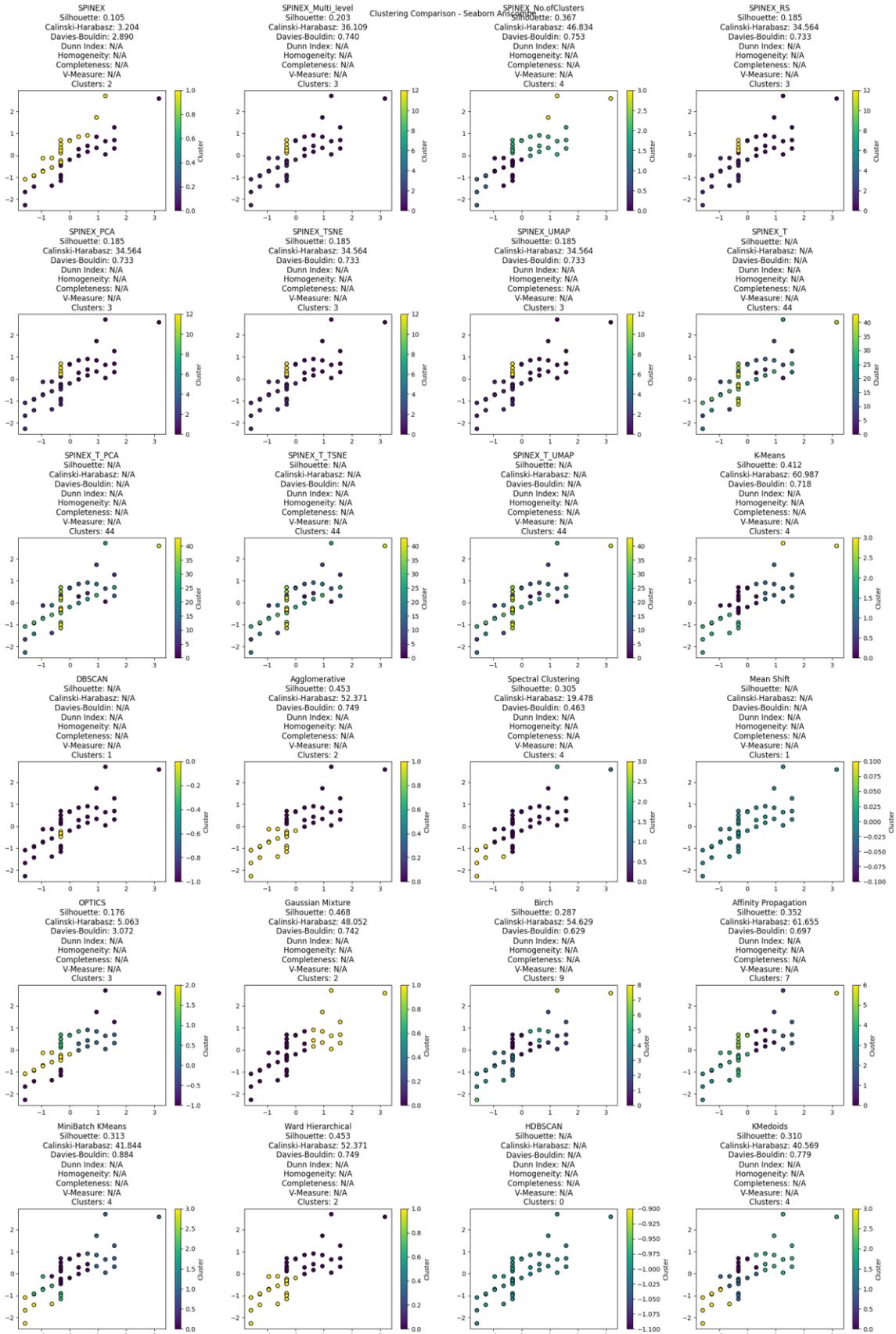

194

54